\documentclass[lettersize,journal]{IEEEtran}
\usepackage{amsmath,amsfonts}
\usepackage{amssymb}
\usepackage{algorithmic}
\usepackage{algorithm}
\usepackage{array}
\usepackage[caption=false,font=small,labelfont=normalfont,textfont=normalfont]{subfig}
\usepackage{textcomp}
\usepackage{stfloats}
\usepackage{url}
\usepackage{verbatim}
\usepackage{graphicx}
\usepackage{cite}
\usepackage{booktabs}
\usepackage{multirow}
\usepackage{color}
\usepackage{makecell}
\usepackage{array}
\usepackage{amssymb}
\usepackage{amsmath}
\usepackage{lineno,hyperref}

\usepackage{tikz}
\usetikzlibrary{arrows.meta, decorations.pathmorphing}
\usepackage[dvipsnames]{xcolor}

\usepackage{etoolbox}

\BeforeBeginEnvironment{IEEEbiography}{\vspace{-12pt}}
\AfterEndEnvironment{IEEEbiography}{\vspace{-8pt}}

\begin{document}

\title{Putting on the Thinking Hats: A Survey on Chain of Thought Fine-tuning from the Perspective of Human Reasoning Mechanism}
\author{Xiaoshu Chen, Sihang Zhou, Ke Liang, Duanyang Yuan, Haoyuan Chen, Xiaoyu Sun, Lingyuan Meng $\text{Xinwang Liu}^*$, \textit{Senior Member, IEEE}
\thanks{Xiaoshu Chen, Ke Liang, Lingyuan Meng, Xiaoyu Sun and Xinwang Liu are with the College of Computer Science and Technology, National University of Defense Technology, Changsha 410073, China (e-mail: xinwangliu@nudt.edu.cn) \par
Sihang Zhou, Duanyang Yuan, and Haoyuan Chen are with the College of Intelligence Science and Technology, National University of Defense Technology, Changsha 410073, China. \par
$*$ Corresponding Author}
}

\markboth{Under Review}%
{Shell \MakeLowercase{\textit{et al.}}: A Sample Article Using IEEEtran.cls for IEEE Journals}


\maketitle

\begin{abstract}
Chain of thought (CoT) fine-tuning aims to endow large language models (LLMs) with reasoning capabilities by training them on curated reasoning traces. It leverages both supervised and reinforced fine-tuning to cultivate human-like reasoning skills in LLMs, including detailed planning, divergent thinking, intuitive judgment, timely reflection, internal thinking, and fact perception, etc. As CoT fine-tuning has advanced, LLMs have demonstrated substantial improvements in tasks such as mathematical reasoning and code generation.
However, existing surveys about CoT fine-tuning primarily focus on technical aspects and overlook a systematic analysis from the perspective of human reasoning mechanisms. Given that the ultimate goal of CoT fine-tuning is to enable LLMs to reason like humans, it is crucial to investigate this technique through the lens of human cognition.
To fill this gap, we present the first comprehensive survey of CoT fine-tuning grounded in human reasoning theory. Specifically, inspired by the well-known Six Thinking Hats framework, which systematically characterizes common human thinking modes using six metaphorical hats, we classify and examine CoT fine-tuning methods through this lens. Furthermore, building upon this theory, we outline potential directions for future research in CoT fine-tuning. In addition, we compile a comprehensive overview of existing datasets and model performances, and a real-time GitHub repository \footnote{https://github.com/AI-Chen/Awesome-CoT-Finetuning} that continuously tracks recent advances in this area is maintained. We hope this survey will serve as a valuable resource to inspire innovation and foster progress in this rapidly evolving field.
\end{abstract}


\section{Introduction}
\setlength{\textfloatsep}{1.25pt} 
 \IEEEPARstart{T}he continuous improvement of reasoning capabilities of large language models (LLMs) \cite{openai2025gpt5, yang2025qwen3technicalreport,kimiteam2025kimik2openagentic,deepseekai2024deepseekv3technicalreport, seed2025tech_seed1_6} has significantly impacted various aspects of daily life. Reasoning, which refers to the cognitive process of drawing conclusions or making judgments based on existing information, knowledge, or logical rules, is a fundamental aspect of human cognition \cite{stenning2012human, evans2019thinking}. Thus, compared to simple language generation or pattern matching, reasoning with LLMs requires the LLMs to establish non-trivial and structured logical pathways between inputs and outputs, which not only reflects the depth of the LLM’s knowledge understanding, but also determines its potential for practical application in complex tasks such as code generation \cite{Anthropic_Claude4_2025}, mathematical reasoning \cite{NEURIPS2024_4b77d5b8}, commonsense question answering \cite{toroghi-etal-2024-verifiable}, and information retrieval \cite{dai2024bias}, etc.

\begin{table}[]
\centering
\scriptsize
\captionsetup{skip=1pt} 
\caption{Comparison between different CoT fine-tuning surveys}
\setlength{\tabcolsep}{0.9mm}
\begin{tabular}{ccccccccccc}
\toprule[1.05pt]
Survey & \cite{2023-towards-reasoning}& \cite{x-of-cot} &  \cite{chu-etal-2024-navigate} & \cite{besta2025reasoning-Blueprint} & \cite{xu2025towardsReinforced-Reasoning} & \cite{chen2025towards-reason-era}  & \cite{system1to2} & \cite{llm-post-training} & \cite{sui2025stop}& Ours \\ \toprule[1.05pt]
Detailed planning             & $\checkmark$  &              & $\checkmark$  &               &               &              &               &              &               & $\checkmark$ \\
Diverse thinking     & $\checkmark$  &              & $\checkmark$  & $\checkmark$  & $\checkmark$  & $\checkmark$ & $\checkmark$  & $\checkmark$ &               & $\checkmark$\\
Intuitive judgment            &               &              &               &               &               &              &               &              & $\checkmark$  & $\checkmark$\\
Timely reflection           & $\checkmark$  & $\checkmark$ & $\checkmark$  & $\checkmark$  & $\checkmark$  & $\checkmark$ & $\checkmark$  & $\checkmark$ &               & $\checkmark$\\ 
Internal thinking    &               &              &               &               &               &              & $\checkmark$  &              & $\checkmark$  & $\checkmark$\\
Fact perception      &               & $\checkmark$ & $\checkmark$  &               & $\checkmark$  & $\checkmark$ & $\checkmark$  &              &               & $\checkmark$\\
\toprule[1.05pt]
\end{tabular}
\label{tab:surveys}
\end{table}

Early LLMs \cite{RLHF, workshop2023bloom176bparameteropenaccessmultilingual} primarily relied on large-scale unsupervised text corpora and instruction-tuning data for language modeling. Their reasoning capabilities largely emerge implicitly \cite{cot, kojima2023largelanguagemodelszeroshot}, without explicit logical supervision. Consequently, they tend to produce answers that appear plausible but are logically incorrect when confronted with tasks requiring rigorous reasoning, thus compromising the reliability and controllability of the output. To address this limitation, Chain of thought fine-tuning (CoT fine-tuning) \cite{zelikman2022star, gsm8k, trung-etal-2024-reft} has emerged as a key technique. It introduces explicit intermediate reasoning steps (i.e., CoT) as supervision signals during training, enabling LLMs to learn ``how to think” rather than simply ``what to answer.” By doing so, it offers multiple benefits, such as enhancing the interpretability of LLMs \cite{mi2025cot}, improving performance on complex tasks \cite{11086374}, and facilitating integration with external feedback \cite{qu2025tool}.

\begin{figure*}[ht]
    \setlength{\belowcaptionskip}{-0.1cm} 
    \centering
    \includegraphics[width=0.85\linewidth]{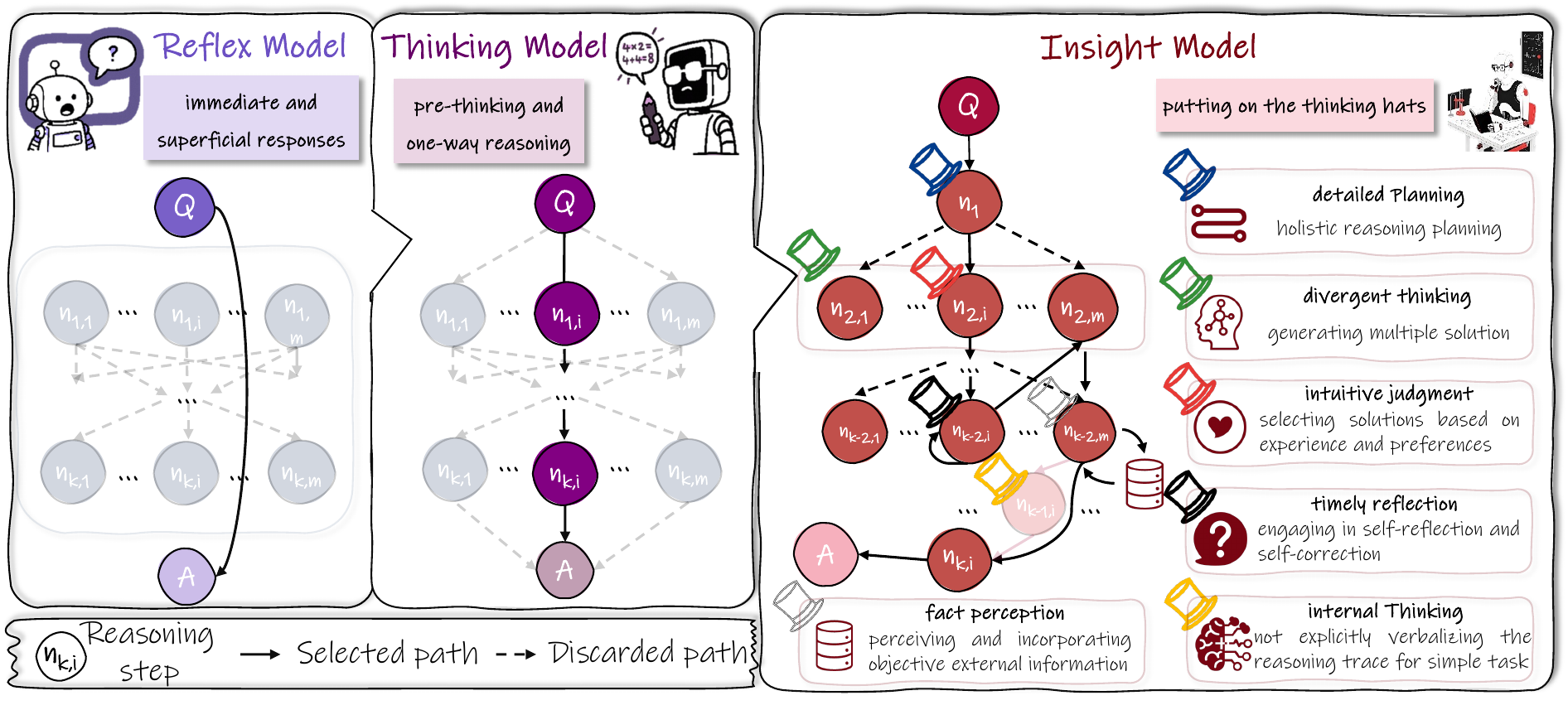}
    \caption{
    The evolution of CoT fine-tuning. It can be divided into two stages: the Thinking Model stage and the Insight Model stage. Before the two stages, LLMs focus on training LLMs to produce direct answers (the Reflex model stage). During the Thinking Model stage, CoT fine-tuning enables LLMs to reason step-by-step. After entering the Insight Model stage, CoT fine-tuning equips LLMs with human-like thinking abilities. Drawing on the Six Thinking Hats framework, we categorize the development of CoT fine-tuning in the Insight Model stage into six distinct dimensions: detailed planning, divergent thinking, intuitive judgment, timely reflection, internal reasoning, and factual perception. These capabilities correspond respectively to the six metaphorical hats: \textcolor{blue}{blue} (management), \textcolor{green}{green} (creativity), \textcolor{red}{red} (intuition), black (criticism), \textcolor{orange}{yellow} (optimism), and \textcolor{gray}{white} (objectivity). Together, these hats are employed interactively to form a comprehensive and mature reasoning framework, as described in the Six Thinking Hats theory.}
    \label{fig:method}
\end{figure*}

As illustrated in Figure \ref{fig:method}, the development of CoT fine-tuning in LLMs can be broadly categorized into two stages: the Thinking Model and the Insight Model stages. Before the Think Model stage, LLMs directly generate final answers from inputs by relying on implicit input–output mappings (the Reflex Model stage). This process primarily depends on pattern matching from training data rather than on genuine reasoning ability. During the Thinking Model stage, researchers introduce one-way CoT supervision to fine-tune LLMs explicitly, enabling them to model the reasoning trajectory from input to output. However, this stage still faces limitations: reliance on fixed, one-way CoTs constrains LLMs to a single logical path, limiting their adaptability to the diverse and dynamic reasoning required by complex tasks. Consequently, LLMs often underperform on tasks characterized by high uncertainty and complexity. To address these challenges, the Insight Model stage introduces advanced techniques to unlock human-like thinking in LLMs, substantially enhancing their reasoning performance in complex scenarios.


With this rapid advance in CoT fine-tuning, several surveys (see Table \ref{tab:surveys}) have attempted to synthesize its research progress. In particular, early surveys \cite{2023-towards-reasoning, x-of-cot, chu-etal-2024-navigate} primarily focused on the role of prompt engineering in enhancing the reasoning capabilities of LLMs and only referenced CoT fine-tuning sporadically, thereby lacking a comprehensive analysis. Although some recent works \cite{besta2025reasoning-Blueprint,xu2025towardsReinforced-Reasoning, chen2025towards-reason-era, system1to2, llm-post-training, sui2025stop} have begun to highlight the pivotal role of CoT fine-tuning in advancing the reasoning abilities of LLMs, most of these reviews remain confined to summarizing technical methods, and still lack in-depth analysis of CoT fine-tuning from the perspectives of human reasoning mechanisms. Therefore, advanced reasoning capabilities that resemble human reasoning, as observed in the Insight Model stage, remain insufficiently explored in existing literature, as summarized in Table \ref{tab:surveys}.

To address this gap, our survey employs the ``Six Thinking Hats" framework \cite{debono1985six} to systematically and comprehensively review the development of human-like reasoning mechanisms enabled by CoT fine-tuning in the Insight Model stage. The Six Thinking Hats describes a classic systematic human thinking mechanism designed to help individuals or teams think in a structured way. Specifically, the framework likens the reasoning process to wearing differently colored ``hats," each representing a distinct mode of thinking, i.e., blue, green, red, black, yellow and white correspond to management, creativity, intuition, criticism, optimism, and objectivity during thinking, respectively. By actively switching between these thinking hats, the framework effectively mitigates biases introduced by linear thinking and enhances the comprehensiveness and depth of problem-solving. As illustrated in Figure \ref{fig:method}, based on the six thinking modes represented by thinking hats, we can classify the developmental trajectories of human-like reasoning mechanisms in the Insight model stage into six key reasoning abilities: detailed planning (management), divergent thinking (creativity), intuitive judgment (intuition), timely reflection (criticism), internalized reasoning (optimism), and factual perception (objectivity). Accordingly, this review proposes a bi-level taxonomy: the top level is grounded in human reasoning modes conceptualized by the Six Thinking Hats, while the base level corresponds to specific techniques that enable these reasoning functions. Utilizing this taxonomy, we comprehensively analyze the role of CoT fine-tuning in cultivating reasoning capabilities in LLMs from dual perspectives: human-like reasoning mechanisms and technical pathways. Notably, this review not only surveys existing technological advances but also offers an in-depth discussion on the potential challenges and future directions of CoT fine-tuning, guided by the Six Thinking Hats framework, thus providing a systematic reference for future research.

\begin{figure*}[ht]
\centering
\begin{tikzpicture}[
  node/.style={
    rectangle,
    draw=red!30!black,
    ultra thin,
    rounded corners=4pt,
    minimum width=1.6cm,
    minimum height=0.75cm,
    align=center,
    fill=red!10,
    font=\scriptsize
  },
  conn/.style={
    ultra thin,
    draw=red!30!black
  }
]

\node[node, text width=2cm, align=center, fill=red!40] (root) at (-2,0) {CoT\\Fine-tuning};

\node[node, text width=1.4cm, align=center, fill=red!30] (insight)   at (-0.1, -0.75) {Insight Model (§\ref{sec: insightmodel})};
\node[node, text width=1.5cm, align=center, fill=red!30] (think) at (-9, -0.75) {Thinking Model (§\ref{sec: thinkingmodel})};

\node[node, text width=1cm, align=center, fill=red!20] (sft)    at (-10, -1.75) {SFT (§\ref{sec: sft})};
\node[node, text width=1cm, align=center, fill=red!20] (rft)    at (-7.8, -1.75) {RFT (§\ref{sec: rft})};

\node[node, text width=0.5cm, align=center, fill=red!20] (blue)    at (-5.5, -1.75) {Blue (§\ref{sec: blue})};
\node[node, text width=0.5cm, align=center, fill=red!20] (green)    at (-3.5, -1.75) {Green (§\ref{sec: green})};
\node[node, text width=0.5cm, align=center, fill=red!20] (red)    at (-1.4, -1.75) {Red (§\ref{sec: red})};
\node[node, text width=0.5cm, align=center, fill=red!20] (black)    at (0.8, -1.75) {Black (§\ref{sec: black})};
\node[node, text width=0.5cm, align=center, fill=red!20] (yellow)    at (2.9, -1.75) {Yellow (§\ref{sec: yellow})};
\node[node, text width=0.5cm, align=center, fill=red!20] (white)    at (5, -1.75) {White (§\ref{sec: white})};

\node[node, text width=1.8cm, align=center] (sft_3)    at (-10, -2.95) {SFT Methods \\(§\ref{sec: sft_1})\\CoT Anotation\\(§\ref{sec: sft_2})};
\node[node, text width=2cm, align=center] (rft_3)    at (-7.8, -2.95) {RFT Methods\\ (§\ref{sec: rft_1})\\Reward Modeling\\ (§\ref{sec: rft_2}};

\node[node, text width=1.8cm, align=center] (blue_3)    at (-5.5, -2.95) {Self
Planning\\ (§\ref{sec: blue_1})\\External Planning\\ (§\ref{sec: blue_2})};
\node[node, text width=1.6cm, align=center] (green_3)    at (-3.5, -2.95) {Self Diversity\\ (§\ref{sec: green_1})\\TTD\\ (§\ref{sec: green_2})};
\node[node, text width=2cm, align=center] (red_3)    at (-1.4, -3.22) {Minimal
-budget\\ (§\ref{sec: red_1})\\Well-structured \\ (§\ref{sec: red_2})\\Safety-oriented\\ (§\ref{sec: red_3}) };
\node[node, text width=1.8cm, align=center] (black_3)    at (0.8, -3.22) {MAD\\ (§\ref{sec: black_1})\\Verifier Feedback\\ (§\ref{sec: black_2})\\Self Reflection\\ (§\ref{sec: black_3})};
\node[node, text width=1.8cm, align=center] (yellow_3)    at (2.9, -3.22) {f \& s
thinking\\ (§\ref{sec: yellow_1})\\Skip Thinking\\ (§\ref{sec: yellow_2})\\Latent Thinking\\ (§\ref{sec: yellow_3})};
\node[node, text width=1.8cm, align=center] (white_3)    at (5, -3.35) {Tool Selection \\ (§\ref{sec: white_1}) \\Tool Calling \\ (§\ref{sec: white_2})\\Response Comprehension\\ (§\ref{sec: white_3})};

\coordinate (out1) at ([yshift=0.3cm]insight.north);
\draw[conn] (root.east) |- (out1.north);
\draw[conn] (out1.south) |- (insight.north);

\coordinate (out2) at ([yshift=0.3cm]think.north);
\draw[conn] (root.west) |- (out2.north);
\draw[conn] (out2.south) |- (think.north);

\coordinate (out10) at ([yshift=0.6cm]rft.north);
\draw[conn] (think.east) |- (out10.north);
\draw[conn] (out10.south) |- (rft.north);

\coordinate (out3) at ([yshift=0.6cm]sft.north);
\draw[conn] (think.west) |- (out3.north);
\draw[conn] (out3.south) |- (sft.north);

\coordinate (out4) at ([yshift=0.6cm]blue.north);
\draw[conn] (insight.west) |- (out4.north);
\draw[conn] (out4.south) |- (blue.north);

\coordinate (out5) at ([yshift=0.6cm]green.north);
\draw[conn] (insight.west) |- (out5.north);
\draw[conn] (out5.south) |- (green.north);

\coordinate (out6) at ([yshift=0.6cm]red.north);
\draw[conn] (insight.west) |- (out6.north);
\draw[conn] (out6.south) |- (red.north);

\coordinate (out7) at ([yshift=0.6cm]black.north);
\draw[conn] (insight.east) |- (out7.north);
\draw[conn] (out7.south) |- (black.north);

\coordinate (out8) at ([yshift=0.6cm]yellow.north);
\draw[conn] (insight.east) |- (out8.north);
\draw[conn] (out8.south) |- (yellow.north);

\coordinate (out9) at ([yshift=0.6cm]white.north);
\draw[conn] (insight.east) |- (out9.north);
\draw[conn] (out9.south) |- (white.north);

\draw[conn] (sft.south) |- (sft_3.north);
\draw[conn] (rft.south) |- (rft_3.north);
\draw[conn] (blue.south) |- (blue_3.north);
\draw[conn] (green.south) |- (green_3.north);
\draw[conn] (red.south) |- (red_3.north);
\draw[conn] (black.south) |- (black_3.north);
\draw[conn] (yellow.south) |- (yellow_3.north);
\draw[conn] (white.south) |- (white_3.north);

\end{tikzpicture}
\caption{Taxonomy of CoT fine-tuning. TTD, MAD, and f\&s thinking refer to test time diversity, multi-agent debate, and the integration of slow and fast
thinking, respectively.}
\label{fig:taxonomy}
\end{figure*}
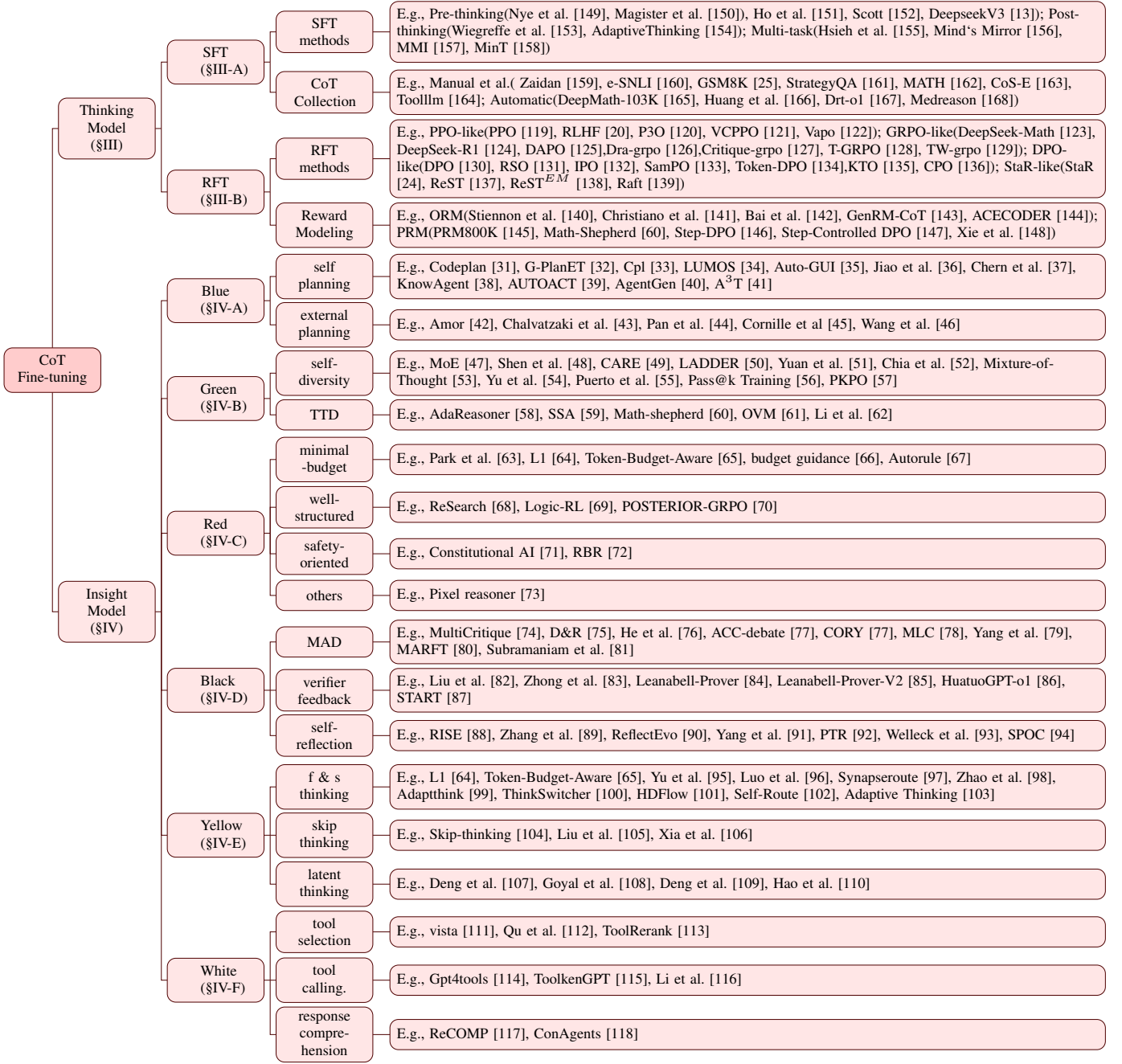

In summary, this survey presents the first comprehensive review of CoT fine-tuning from the perspective of a human systematic thinking framework. The taxonomy is shown in Figure. \ref{fig:taxonomy} and the structure of this survey is organized as follows: First, Section \ref{sec: preliminary} introduces the preliminary of this survey. Section \ref{sec: thinkingmodel} reviews the development of CoT fine-tuning in the Thinking Model stage, systematically tracing the evolution of early CoT fine-tuning methods, and Section \ref{sec: insightmodel} reviews the most representative CoT fine-tuning approaches developed in the Insight Model Stage, based on the proposed bi-level taxonomy framework. Section \ref{sec: datasets} organizes and categorizes commonly used datasets for evaluating reasoning capabilities and provide the performance comparison of representative methods. Section \ref{sec: chanllenge} leverages the Six Thinking Hats framework to analyze key challenges and potential research opportunities in CoT fine-tuning. Finally, Section \ref{sec: conclusion} concludes this survey.

To highlight the contributions of this survey, we summarize the main points as follows:

\textit{1) Comprehensive Review:} We comprehensively investigate typical CoT fine-tuning methods based on a bi-level taxonomy, i.e., top-level (Six Thinking Hats), and base-level (techniques), which offers a novel perspective on surveying CoT fine-tuning.

\textit{2) Insightful Analysis:} Based on the Six Thinking Hats, we analyze the strengths and limitations of existing CoT fine-tuning methods in developing corresponding reasoning abilities of LLMs, which facilitates researchers in selecting an appropriate baseline for their research.

\textit{3) Potential Opportunity:} Building upon the Six Thinking Hats, we discuss the key challenges currently faced by CoT fine-tuning and point out some potential opportunities that will inspire future studies.

\textit{4) Open-source Resource:} A continuously updated GitHub repository is maintained to track the latest developments.

\section{PRELIMINARY}
\label{sec: preliminary}
This section begins with a formal definition of CoT fine-tuning, and then provides a overview of the LLM's six reasoning abilities corresponding to the Six Thinking Hats.

\subsection{Definition of CoT fine-tuning}
The reasoning capability of LLMs refers to their capacity to construct intermediate steps that form a coherent logical mapping bridge between input and output. Specifically, prior to generating an answer, the LLM produces a CoT in the following format:

\begin{equation}
\begin{aligned}
    \mathcal{C} = \{n_k = \{s_k(r_{k-1}), r_k)\}\}_{k=0}^{K}
    \quad s.t. \quad n_k \rightarrow n_{k+1} 
    \label{equation:cot}
\end{aligned}
\end{equation}
where $n_k$ denotes a reasoning logic node that encapsulates both the intermediate reasoning process $s_k(\cdot)$ and reasoning result $r_k$ at the reasoning step $k$, and $\rightarrow$ indicates a single direction connection from one node to the next. In essence, a single CoT captures the reasoning trajectory from the input $Q$ to the final answer $A$, as illustrated in Figure \ref{fig:method}.

CoT fine-tuning is a learnable paradigm that directly optimizes the model's parameters using training data annotated with CoT, thus endowing LLMs with reasoning abilities. In general, in this paper, we define CoT fine-tuning as:
\begin{equation}
    \theta^* = \arg\min_{\theta} \mathbb{E}_{(x, y, c) \sim \mathcal{D}} \left[ \mathcal{L}(f_\theta(x), c \oplus y) \right]
    \label{equation: cot_fine-tuning}
\end{equation}
where $x$, $c$, $y$, and $\oplus$ represent the input question, CoT, answer, and string concatenation, respectively, $\mathcal{L(\cdot)}$ is the loss function, and $f_\theta(x)$ denotes the reasoning result of the generation function with parameters $\theta$.

\subsection{Reasoning Ability Corresponding to Six Thinking Hats}
By applying evolving CoT fine-tuning, LLMs develop human-like reasoning ability defined in the Six Thinking Hats.
\subsubsection{Blue Hat — Detailed Planning} 
The blue hat symbolizes the planning capability that humans employ during the reasoning process. Correspondingly, there are certain nodes in the CoT generated by the LLM that serve as planning nodes. These nodes are responsible for planning the reasoning trace of subsequent global or local reasoning logic.
Formally, let $n_k$ be a planning node whose reasoning result determines the reasoning strategy for subsequent steps. The planning ability can be denoted as:
\begin{equation}
    \exists j>0, n_k \rightarrowtail n_{k+j}
\end{equation}
where $\rightarrowtail$ means the reasoning is constrained by the $n_k$. 

Planning not only provides a structured roadmap for the entire CoT but also effectively prevents unnecessary exploration or detours, avoiding redundant computations.

\subsubsection{Green Hat — Divergent Thinking} 
The green hat represents divergent thinking, characterized by exploring multiple solutions to a single problem. Specifically, for the reasoning process of LLMs, it often explores multiple possible solutions from a single reasoning node to subsequent ones, i.e.,
\begin{equation}
    \exists m\geq0, n_{k} \rightarrow \{n_{k+1,i}\}_{i=0}^{m}
\end{equation}
It should be noted that for multiple solutions, a verifier or reward model is usually used to select the most feasible reasoning solution as the optimal one to improve the feasibility efficiency of subsequent reasoning.

Divergent thinking is indispensable in facilitating a) enhancing problem-solving capabilities and b) strengthening the robustness of reasoning.

\subsubsection{Red Hat — Intuitive judgment}
The red hat indicates the intuitive judgment employed by humans during the reasoning. In many real-world scenarios, problem-solving often lacks sufficient information to clearly determine the superiority of one solution over others. In such cases, humans tend to rely on intuition to prioritize solutions that align more closely with their prior knowledge or subjective preferences.
Similarly, when reasoning with LLMs, it is sometimes infeasible to employ a verifier, or their effectiveness may be limited. In such cases, the LLM is also able to make decisions like humans based on its internal knowledge and learned preferences. This intuition-driven decision-making process can be formalized as:
\begin{equation}
    n_{k+1}^* = \arg\max_{m}(\sum_{i=0}^{N}f_i(n_{k+1,m}))
\end{equation}
where $f_i(\cdot)$ is the score function applied during judgment.

Intuitive judgment is vital in reasoning with LLMs, especially in scenarios characterized by incomplete information, time constraints, or high complexity.

\subsubsection{Black Hat — Timely Reflection}
The black hat symbolizes the reflective capability inherent in human reasoning. When facing complex problems, reasoning errors are often inevitable. Therefore, the ability to promptly reflect on and correct errors is essential. This timely reflective capability has also been incorporated into LLM reasoning via CoT fine-tuning, which is formulated as:
\begin{equation}
    \exists j \geq 0, n_{k+j}^* = n_{k, m}
\end{equation}
i.e., after reflecting on the reasoning path reaching $n_{k+j-1}$, the LLM may revisit the previously erroneous node in the reasoning path and proceed with an alternative, more accurate reasoning path informed by the reflection.

This timely reflection, which involves actively retracing and selecting a more optimal reasoning path, makes the LLM reasoning no longer constrained to a forward-only direction, and significantly improves overall robustness.

\subsubsection{Yellow Hat — Internal Thinking}
The yellow hat represents confidence and optimism in the reasoning. Concretely, when faced with relatively simple problems, humans are usually so confident in their reasoning that they do not even explicitly record their reasoning process but think about it in their mind. CoT fine-tuning also focuses on developing this ability of internalized thinking. Formally, internalized thinking can be expressed as:
\begin{equation}
    \exists 0 \leq k \leq K, \exists J \geq 0,  \{n_{k+j}= none\}_{j=0}^{J}
\end{equation}
In other words, there are some successive reasoning steps not being explicitly output when LLMs are confident.

This form of optimistic reasoning enables LLMs to allocate more attention to complex reasoning subtasks within the current question. Furthermore, by reducing the need for explicit intermediate outputs, the reasoning efficiency of LLMs is significantly improved.

\subsubsection{White Hat — Fact perception}
The white hat refers to the model's capacity for factual perception during reasoning. Given the limitations of human knowledge, individuals often rely on tools to perceive external objective facts during reasoning. Similarly, LLMs also face knowledge boundaries. Thus, enhancing their reasoning capabilities of utilizing external tools for factual perception through CoT fine-tuning has become a key research direction. The mechanism of factual perception within a CoT can be formally expressed as follows:
\begin{equation}
    s(r_{k-1}, tool)
\end{equation}
which means that LLMs can apply tools to perceive object facts during reasoning.

Overall, the continuous perception of objective facts ensures hallucinated reasoning is mitigated and enhances the LLM’s adaptability to dynamic environments.

\section{Being a Thinking Model}
\label{sec: thinkingmodel}

\begin{figure*}[ht]
    \centering
    \includegraphics[width=0.88\linewidth]{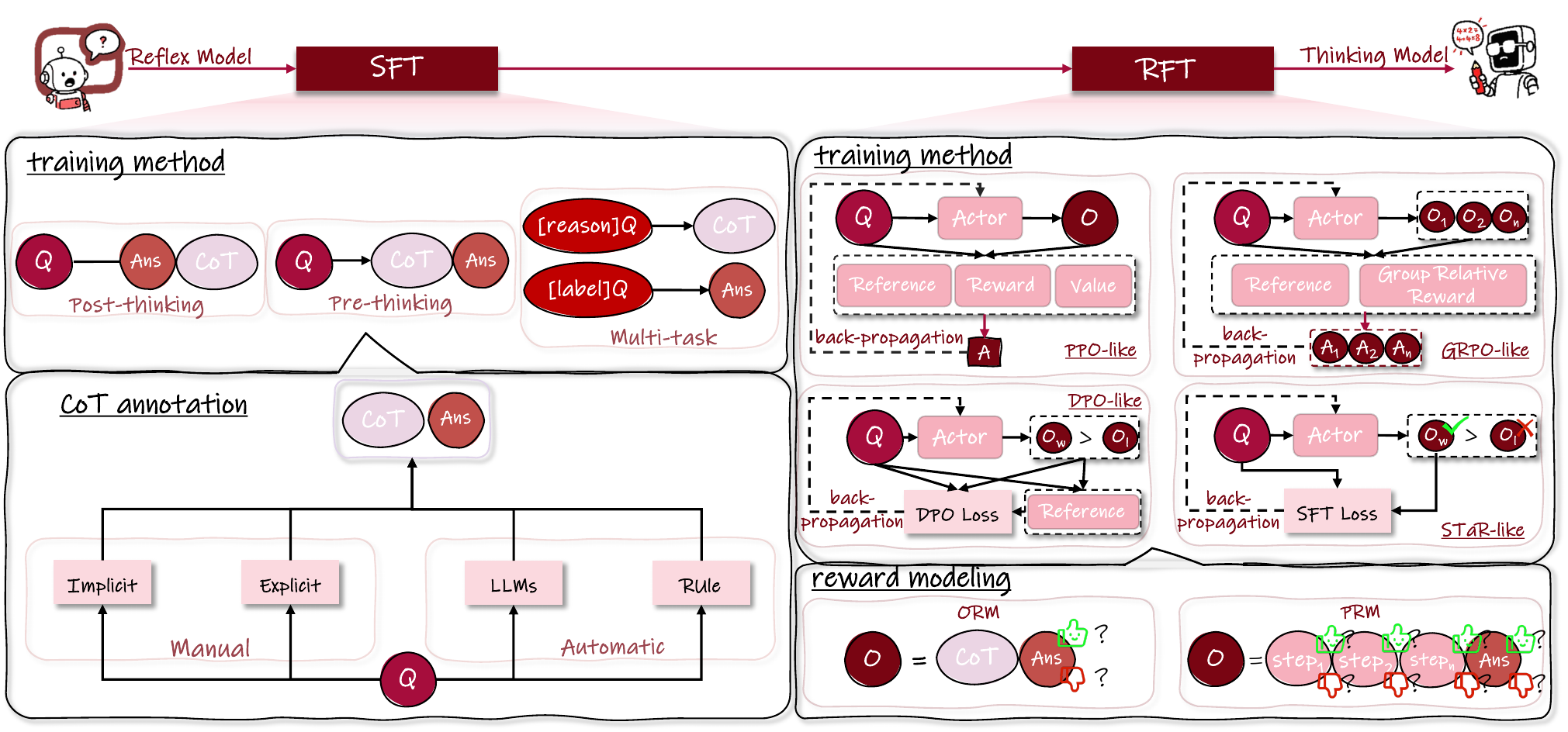}
    \caption{The training pipeline of converting LLMs from the Reflex Model to the Thinking Model. The primary objective of SFT is to train LLMs to generate both the CoT and the final answer given an input question, while that of RFT is to strengthen the reasoning ability of LLMs.}
    \label{fig: bethinker}
\end{figure*}

Becoming a competent Thinking Model forms the foundation for evolving into an advanced Insight Model. Therefore, before summarizing the Insight Model, this section introduces the core techniques for transforming LLMs from the Reflex Model into the Thinking Model. Specifically, as shown in Figure \ref{fig: bethinker}, this transformation typically involves two stages: supervised fine-tuning (SFT) and reinforced fine-tuning (RFT). Accordingly, this section is organized into two parts, each discussing the training methods of SFT and RFT, respectively.


\subsection{Supervised Fine-tuning}
\label{sec: sft}
\subsubsection{Training Methods}
\label{sec: sft_1}
The primary distinction across different SFT approaches lies in the design and modification of training objectives. Broadly, these training objectives can be categorized into three types: pre-thinking \cite{nye2021workscratchpad, magister-teaching, ho-etal-2023-teacher, wang-etal-2023-scott,fu-teeacher, li-etal-2024-mode-cotd}, post-thinking \cite{wiegreffe-etal-2021-pos-hoc, chen2024distilling, zhang2026answer}, and multi-task learning \cite{hsieh-etal-2023-Step-by-Step, liu-etal-2024-minds, chen-etal-2024-learning-MMI, liang-etal-2024-mint, wang2025qcrd, yang2026morepair}.


Among them, \textbf{pre-thinking} is the most widely adopted. Its training objective can be formalized as
\begin{equation}
        \theta_{llm}^* = \arg\min_{\theta_{llm}} \mathbb{E}_{(x, y, c) \sim \mathcal{D}} \left[ l_{ce}(f_{\theta_{llm}}(x), c \oplus y) \right]
\label{eq: prethinking}
\end{equation}
where $\oplus$ represents the string concatenation, $l_{ce}$ is the cross entropy loss function and $\theta_{llm}$ denotes the parameters of the LLM. This approach enables the model to generate step-by-step reasoning before producing the final answer. 

In contrast, \textbf{post-thinking} trains the LLMs to first output the answer and then provide the explanation, i.e., converting $c \oplus y$ to $y \oplus c$ in Equation \ref{eq: prethinking}.

\textbf{Multi-task learning} introduces specific prefix instructions to control whether the model generates the reasoning process or the final answer, depending on the task requirements. And we can denote its training object as
\begin{equation}
    \begin{split}
        \theta_{llm}^* = \arg\min_{\theta_{llm}} \mathbb{E}_{(x, y, c) \sim \mathcal{D}} [ l_{ce}(f_{\theta_{llm}}([reason] \oplus x), c) \\ + l_{ce}(f_{\theta_{llm}}([label] \oplus x), y)]
    \end{split}
\end{equation}
where $[reason]$ and $[label]$ are the prefix instructions.

Overall, pre-thinking is the most widely adopted due to its “reasoning-before-answering” nature, which inherently enables the LLMs to decompose complex problems into simpler sub-tasks and integrate their solutions. Therefore, most existing LLMs commonly adopt the pre-thinking paradigm during reasoning \cite{guo2025deepseekr1, comanici2025gemini, dubey2024llama3}. 
However, pre-thinking faces two major challenges. The first is the unfaithfulness problem \cite{lyu-etal-2023-faithfulcot, lanham2023measuring-faithfulness}, where the conclusion of the reasoning process is inconsistent with the final answer. Some studies have attempted to address this issue—for example, Scott \cite{wang-etal-2023-scott}, FRODO \cite{paul-etal-2024-Faithfulness-dpo} and CST\cite{hase2026counterfactual} adopt a counterfactual-inspired approach to enforce the model’s final answer to strictly follow the CoT. The second is the error accumulation, where minor mistakes in early reasoning steps gradually amplify along the CoT, eventually leading to incorrect answers \cite{chen2024distilling}. This also reflects a key distinction between Thinking Models and Insight Models: the latter incorporate self-correction capabilities to correct the early errors.

In contrast, post-thinking allows LLMs to produce answers directly, resulting in faster response speed. Since the CoT is generated after the answer, this approach also mitigates the error accumulation \cite{chen2024distilling}. However, because the answer is generated prior to the CoT, post-thinking lacks the problem-decomposition ability that pre-thinking offers. Overall, post-thinking is particularly suitable for transforming small language models into Thinking Models, as they are typically applied in simple reasoning scenarios where faster responses are prioritized and reasoning errors in CoT are more frequent.

Regarding multi-task learning, it also benefits from rapid answer generation, and prior studies have shown that it often yields more stable training compared to post-thinking. Nevertheless, because reasoning and answer generation occur in separate sequences, the unfaithfulness problem arises more frequently than pre-thinking, which reduces the reliability of the reasoning process and constrains the broader applicability of multi-task learning. Recently, researchers have proposed using mutual information to migrate the unfaithfulness problem \cite{chen-etal-2024-learning-MMI}.

\subsubsection{Acquiring CoT}
\label{sec: sft_2}
The foundation of the aforementioned SFT is the availability of sufficient CoT data to serve as training labels. In this regard, we conduct a comprehensive survey of relevant studies and categorize existing approaches for acquiring CoT data into two main categories—manual and automatic annotation, as illustrated in Figure \ref{fig: bethinker}.

\textbf{Manual Annotation.}
Manual acquisition can be further divided into two methods: implicit and explicit annotation. The former refers to reasoning traces that emerge as by-products when annotators provide the final answers, whereas the latter requires annotators to explicitly articulate their reasoning process in natural language.

For implicit annotation, Zaidan et al.\cite{AnnotatorRationales} first ask annotators to highlight spans of the input text that supported their labeling decisions. Building on this idea, datasets such as e-SNLI \cite{e-SNLI} and its multi-model version e-SNLI-VE \cite{do2020esnli-ve}, MultiRC \cite{MultiRC2018}, FEVER \cite{thorne-etal-2018-fever}, and CoS-E\cite{CoS-E} further require annotators to highlight words they considered essential for natural language inference, reading comprehension, fact verification and commonsense reasoning. Beyond span annotation, WebGPT \cite{nakano2021webgpt} records annotators’ browser interaction sequences during information retrieval as implicit CoT, which is shown to enhance document retrieval performance.

Explicit annotation requires more effort to annotate CoT, but it enables LLMs to perform better. Recently, an increasing number of datasets incorporate carefully crafted CoT for tasks such as mathematical reasoning \cite{gsm8k, math}, commonsense reasoning \cite{CoS-E}, tool use \cite{toolllm}, domain-specific question answering \cite{JAMA, zuo2025medxpertqa, rein2024gpqa}, and natural language inference \cite{e-SNLI}.

\textbf{Automatic Annotation.}
Since manual CoT annotation is considerably more costly than directly labeling answers, a large body of research has explored how to generate CoT data automatically. In general, automatic methods can also be divided into two subcategories: LLM-based and rule-based.

The feasibility of LLM-based automatic annotation primarily arises from the emergent abilities of LLMs. LLMs with over 100B parameters can, when provided with a few CoT examples, generate useful CoT that enhance problem solving \cite{cot}. Accordingly, many studies \cite{magister-teaching, ho-etal-2023-teacher, wang-etal-2023-scott,fu-teeacher, li-etal-2024-mode-cotd, zhu-etal-2024-pad, zhu-math-techer,hsieh-etal-2023-Step-by-Step, liu-etal-2024-minds} directly prompt advanced LLMs to produce CoT for training smaller LLMs. Qin et al. \cite{qin2024o1} and Huang et al. \cite{huang2024o1} show that weaker LLMs trained on these CoTs can match or surpass the performance of their teachers. DeepMath-103K \cite{he2025deepmath} constructs a large-scale mathematical reasoning dataset using LLM-generated CoTs. LLaVA-o1 \cite{xu2024llava}, Marco-o1 \cite{zhao2024marco}, and Mulberry \cite{yao2024mulberry} further demonstrate the effectiveness of LLM-generated CoT for multimodal reasoning. Moreover, post-processing methods such as tree search \cite{huang2024o1} and multi-agent debate \cite{wang2024drt} are often applied to improve CoT quality. However, such CoTs frequently contain noise. To mitigate this, researchers employ external verifiers to filter incorrect instances. A common and general solution is the answer verifier \cite{fu-teeacher, hsieh-etal-2023-Step-by-Step,chen2024distilling}, which discards CoT whose final answer conflicts with ground truth. Task-specific verifiers have also been proposed: code compilers can check syntactic correctness in code reasoning \cite{zhu-etal-2024-pad}, while external knowledge bases verify factual accuracy in commonsense reasoning \cite{cok}.

Beyond LLM-based automatic generation, rule-based approaches offer an alternative research direction. For instance, MedReason \cite{MedReason} converts subgraph search results from a structured medical knowledge graph into CoT. However, designing similar methods in other domains remains highly challenging.

\begin{table}[]
\caption{Comparison of three SFT methods. \textit{max scale} denotes the largest model scale on which each method has been validated.}
\centering
\scriptsize
\renewcommand\arraystretch{1}
\setlength{\tabcolsep}{1.0mm}
\begin{tabular}{cccc}
\hline
                       & \textbf{max scale} & \textbf{targeted scenarios} & \textbf{reasoning speed} \\ \hline
\textbf{pre-thinking}  & \textgreater 100B       & high task complexity        & slow                     \\
\textbf{post-thinking} & \textless 3B            & low task complexity         & fast                     \\
\textbf{multi-task}    & \textless 3B            & low task complexity         & fast                     \\ \hline
\end{tabular}
\label{tab: sft_three}
\end{table}

\subsubsection{Takeaways}
\begin{itemize}
    \item There are three main SFT paradigms for converting the Reflecive Model into the Thinking model. Their differences are summarized in Table \ref{tab: sft_three}.
    \item There remains substantial room for improvement across these methods, particularly in the widely adopted pre-thinking, where the unfaithfulness problem \cite{tutek-etal-2025-measuring} and error accumulation severely limit the practical utility of CoT.
    \item Annotating CoT serves as the foundation for all three paradigms. Manual annotation depends on domain experts, making it both costly and time-consuming. Automatic generation leverages LLMs or rule-based pipelines to efficiently produce CoT data. However, the resulting CoT inadvertently inherits biases from the LLMs.  
\end{itemize}

\subsection{Reinforced Fine-tuning}
\label{sec: rft}

\subsubsection{Training Methods}
\label{sec: rft_1}
RFT can further enhance the reasoning performance of LLMs that have already undergone SFT. In descending order of resource requirements, methods in RFT can be categorized into four types: PPO-like, GRPO-like, DPO-like, and STaR-like. In the following, we first provide background knowledge on RFT and then introduce these four types of methods in turn.

\textbf{Background of RFT.} Existing RFT for LLMs are fundamentally rooted in the early developments of reinforcement learning \cite{REINFORCE, AC, TRPO, ACER, ACKER, A2C, A3C}, ranging from the REINFORCE \cite{REINFORCE} to the A3C \cite{A3C}. The general objective in these reinforcement learning algorithms is to maximize:
\begin{equation}
    \mathcal{J}(\theta_{rft}) = \mathbb{E}_{x \in P(X), o \in \pi_{rft}(o|x)}\textbf{[}\frac{1}{|o|}\sum_{t=1}^{|o|}\pi_{rft}(o_t|q,o_{<t})A_t\textbf{]}
    \label{equation: rft}
\end{equation}
Here, $o = r \oplus y$ denotes the token sequence generated by the LLM $\pi_{rft}$ with parameters $\theta_{rft}$ (initialized from $\pi_{sft}$ trained by SFT), which is also referred to as the actor model, for input $x$. $A_t$ represents the advantage of $o_t$, which measures how much better the current output of the actor model is compared to the expected output. The specific forms for computing $A_t$ will be elaborated in the subsequent specific RFT algorithms. In general, the meaning of Eq. \ref{equation: rft} lies in guiding the actor model to generate solutions with larger advantages.

\textbf{PPO-like RFT.}
Proximal Policy Optimization (PPO) \cite{PPO} is originally proposed by Schulman et al., and later gains prominence through reinforcement learning with human feedback (RLHF) \cite{RLHF}. 

In RLHF, it first trains a reward model, which is responsible for computing the reward value of $o$, reflecting the extent to which $o$ aligns with the desired output. The loss function for training the reward model is defined as follows:
\begin{equation}
    \mathcal{L}(\theta_{rm}) = -log(\sigma(rm(x,o_w) - rm(x, o_l)))
\end{equation}
where $o_w$ and $o_l$ are two different outputs of $\pi_{sft}$ for input $x$, with $o_w$ being the one that better conforms to the desired standard compared to $o_l$, and $rm(\cdot,\cdot)$ is the output scalar of the reward model.

Once the reward model has been trained, the reward $r_t$ for $o_t$ can be defined as follows:
\begin{gather}
r_t = \begin{cases}
log(\frac{\pi_{ref}(o_t|x, o<t))}{\pi_{rft}(o_t|x, o<t)}  & t < |o| \\
log(\frac{\pi_{ref}(o_t|x, o<t))}{\pi_{rft}(o_t|x, o<t)} + rm(x, o)  & t = |0|
\end{cases}
\label{reward}
\end{gather}
where $\pi_{ref}$ is a frozen copy of $\pi_{sft}$ before undergoing RFT, $\frac{\pi_{ref}(o_t|x, o<t))}{\pi_{sft}(o_t|x, o<t)}$ represents a penalty that prevents the output distribution of $\pi_{rft}$ during RFT from deviating too far from that of $\pi_{ref}$. The key insight of $r_t$ is that RLHF encourages $\pi_{rft}$ to be rewarded only when $rm(x, o)$ is relatively large; otherwise, $\pi_{rft}$ is expected to remain consistent with the outputs of $\pi_{sft}$. Once $r_t$ is obtained, A simple formulation for calculating $A_t$, which is often optimized with Generalized Advantage Estimation (GAE) \cite{schulman2018highdimensionalcontinuouscontrolusing}, is given by:
\begin{equation}
    A_t = \sum_{t^{'}>t}^{|o|}\gamma^{t^{'}-t}r_{t^{'}} - V(o_{\le t})
\end{equation}
where $\gamma \in [0,1]$ denotes the discount factor, by which assigns greater importance to immediate rewards than to future rewards, and $V(o_{\le t})$ represents the expected return estimated by the value model parameterized by $\theta_v$ for the partial output $o_{\le t}$ produced by $\pi_{rft}$. The value model is also typically initialized from $\pi_{sft}$ and trained with the following loss function to estimate the expected future reward:
\begin{equation}
    \mathcal{L}(\theta_{v}) = \sum_{t=0}^{|o|}(\sum_{t^{'}>t}^{|o|}\gamma^{t^{'}-t}r_{t^{'}} - V(o_{\le t}))^2
\end{equation}
Finally, in the context of RLHF, Equation \ref{equation: rft} can be expressed as

\begin{equation}
    \begin{split}
    \mathcal{J}(\theta_{rft}) = \mathbb{E}_{x \in P(x), o \in \pi_{old}(o|x)}\textbf{[}\frac{1}{|o|}\sum_{t=1}^{|o|} min[\frac{\pi_{rft}(o_t|q,o_{<t})}{\pi_{old}(o_t|q,o_{<t})}A_t, \\ 
    clip(\frac{\pi_{rft}(o_t|q,o_{<t})}{\pi_{old}(o_t|q,o_{<t})}, 1-\epsilon, 1+\epsilon)A_t]  \textbf{]}
    \end{split}
\label{equation:ppo}
\end{equation}
where $\pi_{old}(o_t \mid q, o_{<t})$ denotes the probability distribution of generating $o_t$ under the actor model from $N$ iterations prior to the current training step, and $\epsilon$ is a clipping-related hyper-parameter. Compared to Eq. \ref{equation: rft}, the clipping strategy in Equation \ref{equation:ppo} prevents overly parameter updates of the actor model during RFT, thereby enhancing training stability.

Although RLHF performs well in improving LLMs' reasoning, it suffers from limitations such as instability during training and heavy reliance on the quality of the reward model. To address these issues, several works proposed improvements. For instance, 
P3O \cite{P3O} introduces a penalty term into the objective function to ensure smoother gradient updates; and VC-PPO \cite{VCPPO} and VAPO \cite{yue2025vapo} decouples the computation of GAE between the actor and value models, thereby mitigating the decay of reward signals and improving LLMs’ performance in long CoT reasoning scenarios. Nevertheless, these approaches still require maintaining at least four models—the actor, reference, reward, and value models—during RFT, which leads to excessive training resource consumption. This issue remains an urgent challenge to be solved.

\textbf{GRPO-like RFT.} GRPO-like RFT originates from DeepSeek-Math \cite{he2025deepmath}, which demonstrated that the relative reward of a group of sampled LLM outputs for the same input $x$ can serve as an estimator of the expected advantage, thereby eliminating the need for a value model in RFT. Specifically, in DeepSeek-Math, Equation \ref{equation: rft} can be rewritten as:
\begin{equation}
    \begin{split}
        \mathcal{J}(\theta_{rft}) & = \mathbb{E}_{x \in P(x), \{o^i\}_{i=1}^{G} \in \pi_{old}(O|x)} \frac{1}{G}\sum_{i=1}^{G}\frac{1}{|o|}\sum_{t=1}^{|o|} \\
         & \textbf{\{} min[\frac{\pi_{rft}(o^i_t|q,o^i_{<t})}  {\pi_{old}(o^i_t|q,o^i_{<t})}A^i_t, clip(\frac{\pi_{rft}(o^i_t|q,o^i_{<t})}{\pi_{old}(o^i_t|q,o^i_{<t})}, \\
         & 1-\epsilon, 1+\epsilon)A^i_t]
          + \beta \mathbb{D}_{kl}[\pi_{rft}||\pi_{ref}] \textbf{\}}
    \end{split}
\end{equation}
where $A^i_t$ can be formalized as follows:
\begin{equation}
    A^i_t = \frac{r^i_t - mean(\{rm(x, o^i)\}_{i=1}^G{})}{std(\{rm(x, o^i)\}_{i=1}^G{})},
\end{equation}
$\mathbb{D}_{kl}[\pi_{rft}||\pi_{ref}]$ serves as a KL-divergence penalty to prevent $\pi_{rft}$ from producing outputs that diverge excessively from those of $\pi_{ref}$, and $\beta$ controls the strength of the penalty.

Building on DeepSeek-Math, DAPO \cite{yu2025dapo} introduces decoupled clip and dynamic sampling policy optimization to alleviate issues such as entropy collapse and reward noise. GSPO \cite{zheng2025GSPO} shifts the optimization granularity of GRPO from the token level to the sequence level, thereby improving training stability when applying GRPO within the MoE \cite{MoE}. DRA-GRPO \cite{Dra-grpo} incorporates semantic diversity of reasoning paths into the reward signal, thereby encouraging the model to explore different reasoning trajectories. Critique-GRPO \cite{Critique-grpo} combines natural language critiques with numerical feedback to further enhance the reasoning performance of LLMs. RL-ZVP \cite{RL-ZVP} and ExGRPO \cite{ExGRPO} mitigate the low data utilization issue in GRPO by employing an entropy-guided advantage formulation and an experience replay mechanism, respectively. Moreover, approaches such as DiffuCoder \cite{gong2025diffucoder}, T-GRPO \cite{T-GRPO}, VideoChat-R1 \cite{Videochat-r1}, and TW-GRPO \cite{TW-grpo} extend the application of GRPO to scenarios including masked diffusion models and multimodal reasoning.

\textbf{DPO-like RFT.} Direct Preference Optimization (DPO) \cite{dpo} further eliminates the need for a reward model in RFT by regarding $\pi_{rft}$ itself as a reward model. Concretely, DPO defines its optimization goal as minimizing:
\begin{equation}
    \begin{split}
        \mathcal{J}(\theta_{rft})  = \mathbb{E}_{x\in P(X), (o_w, o_l) \in \pi_{sft}(o|x)} [\sigma(\beta log\frac{\pi_{rft}(o_w|x)}{\pi_{ref}(o_l|x)} \\  - \beta log\frac{\pi_{rft}(o_l|x)}{\pi_{ref}(o_l|x)})]
    \end{split}
\end{equation}
Essentially, this objective encourages the LLM to adjust its output distribution to favor the preferred response $o_w$ over the less desirable one $o_l$.

The follow-up methods of DPO can be categorized based on their primary contributions: loss-based modifications, supervision granularity and model simplification. For loss-based modifications, RSO \cite{rso} replaces the sigmoid loss with a hinge loss to improve training dynamics, while IPO \cite{IPO} introduces a regularization term in the training objective to mitigate overfitting. SamPO \cite{sampo} downsamples an equal number of features from positive and negative preferences to eliminate length bias. For supervision granularity, Token-level DPO\cite{token-dpo,RTO} refines the preference supervision signal from the sequence level down to the token level, and KTO\cite{KTO} leverages datasets labeled with binary signals to train LLMs, improving DPO’s performance in scenarios with imbalanced data. As for model simplification, OAIF \cite{OAIF} demonstrates the feasibility of converting DPO into an on-policy RFT framework, and CPO \cite{CPO} further eliminates the need for a reference model, validating its effectiveness on machine translation tasks.

\textbf{STaR-like RFT.}
STaR-like RFT \cite{zelikman2022star, ReST, singh2024beyond, DBLP:journals/tmlr/Dong0GZCPDZS023, V-STaR, yang-etal-2024-weak, dou-etal-2024-rest} is essentially a self-training approach, where $\pi_{ref}$ is discarded. Similar to DPO-like RFT, it requires collecting multiple reasoning trajectories and conclusions generated by $\pi_{rft}$ for the same input problem, and then partitioning them into correct and incorrect sets using strategies such as answer matching or self-consistency checks. The difference is that STaR-like RFT fine-tunes the LLM only on the correct set using the same objective as in the SFT stage, thereby steering the distribution of $\pi_{rft}$ toward correct reasoning trajectories.

\subsubsection{Reward modeling with Preference Data}
\label{sec: rft_2}
Regardless of the specific RFT method, collecting preference data $(o_w, o_l)$ and conducting reward modeling is an indispensable step. Therefore, we conduct a systematic investigation of this process. In general, reward modeling with preference data can be categorized into outcome rewards modeling (ORM) and process rewards modeling (PRM), which correspond to different reward granularity of the objective function in RFT, i.e., the ORM provides reward signals only for the correctness of the final outcome $y$ in the complete output $o$, whereas the PRM provides rewards for the correctness of each reasoning step $c$ within $o$, and even for individual token-level reasoning.

\textbf{Outcome Rewards Modeling.}
ORMs are typically categorized into three modeling approaches: specialized-model-based, rule-based, and LLM-based reasoning.

The specialized-model-based approach is the most common and has been widely adopted in PPO-like methods. Specifically, it trains a discriminative LLM on human-labeled preference pairs $(o_w, o_l)$ to output a scalar reward for the reasoning results of $\pi_{rft}$\cite{RLHF, summarizeRLHF, Deepreinforcement, harmlessassistantRL, Fine-grainedhumanfeedback, gemini1, Qwen1, Kimik1.5}. Some studies argue that discriminative reward models are prone to reward hacking \cite{rewardhacking1, rewardhacking2, rewardhacking3} and have therefore explored generative reward models. For instance, GenRM-CoT \cite{genrmcot} first generates a scored reasoning process, then produces a judgment of $\pi_{rft}$’s reasoning accuracy, which is finally used as the reward signal.

Since training specialized models requires costly human annotations for preference data, subsequent rule-based and LLM-based approaches explore ways to reduce annotation overhead. Rule-based methods typically assume that a correct final result implies a correct reasoning process. Under this assumption, they directly check whether sampled outputs from $\pi_{rft}$ match the reference answers from SFT data, thus assigning a reward of 1 if matched and 0 otherwise \cite{guo2025deepseekr1, logicrl, acecoder, qwen2025qwq32b, Skywork-o1, acemath, trung-etal-2024-reft}. In scenarios like DPO-like approaches or training specialized reward models, rule-based methods can also directly partition sampled outputs into the corresponding $o_w$ and $o_l$. However, rule-based methods require explicit reference answers, making them applicable mainly to tasks such as mathematics \cite{guo2025deepseekr1, logicrl} and programming \cite{acecoder}, while limiting their generalizability.

Consequently, LLM-based reasoning methods become useful in broader scenarios. These methods prompt $\pi_{rft}$ to self-evaluate \cite{Self-rewarding} or use another LLM \cite{constitutionalai, lee2023rlaif, d-rlaif, LLM-as-a-judge} for evaluation, producing an accuracy score as the reward or splitting the sampled outputs. By avoiding reliance on explicit reference answers, LLM-based reasoning methods apply to more general reasoning tasks, though their reward reliability is lower than that of rule-based methods.

\textbf{Process Rewards Modeling.}
A correct final answer does not guarantee the correctness of the underlying reasoning process. Moreover, even when the final conclusion is wrong, not every intermediate step is necessarily inferior to the corresponding step in a correct derivation. Consequently, outcome-level rewards may mislead $\pi_{rft}$. To mitigate this issue, PRMs have been proposed. The development of PRMs mirrors that of ORMs. Early approaches trained specialized reward models on human-annotated, step-level preference data. For example, Lightman et al. released PRM800K \cite{prm800}, which provides human-annotated step-level preferences and was used to train reward models that deliver step-level signals for $\pi_{rft}$. Subsequent work followed the ORM trajectory, developing rule-based \cite{Step-DPO, Math-Shepherd, Step-ControlledDPO, Rest-mcts*} and LLM-based reasoning approaches \cite{Rest-mcts*, mcts1, mcts2, mcts3}. For instance, Math-Shepherd \cite{Math-Shepherd} uses, from the current step, the proportion of sampled rollouts that eventually reach the correct answer as the reward signal. LLM-based reasoning is often combined with Monte Carlo Tree Search (MCTS) to efficiently obtain step-level feedback in PRMs \cite{mcts1, mcts2, mcts3}. Building on these advances, some methods \cite{token-dpo} refine the feedback granularity to the token level, precisely attributing the contribution of each token. 
Beyond the aforementioned approaches, some works \cite{lu2024autopsv, freeprocessrewardsprocess, ovm} investigate implicitly distributing the overall reward signal from an outcome reward model across individual reasoning steps. This allows the outcome reward model to function as an implicit process reward model, thereby eliminating the need for explicit step-level supervision.

Overall, PRM provides finer-grained feedback than ORM, theoretically enabling a higher performance ceiling. However, obtaining fine-grained feedback incurs higher costs compared to coarse-grained outcome-level signals. On the other hand, the empirical success of ORM-based methods such as DeepSeek-R1 \cite{guo2025deepseekr1} demonstrates that PRM is more susceptible to reward hacking \cite{amodei2016concreteproblemsaisafety} compared to ORM. By contrast, ORM provides a simpler and more direct reward signal, making it less vulnerable to reward hacking and leading to more stable training in large-scale training. Thus, a promising research direction lies in mitigating reward hacking in PRM \cite{cheng2025stopsummationminformcredit, yin-etal-2025-dynamic}.

\begin{figure}[]
    \centering
    \includegraphics[width=0.95\linewidth]{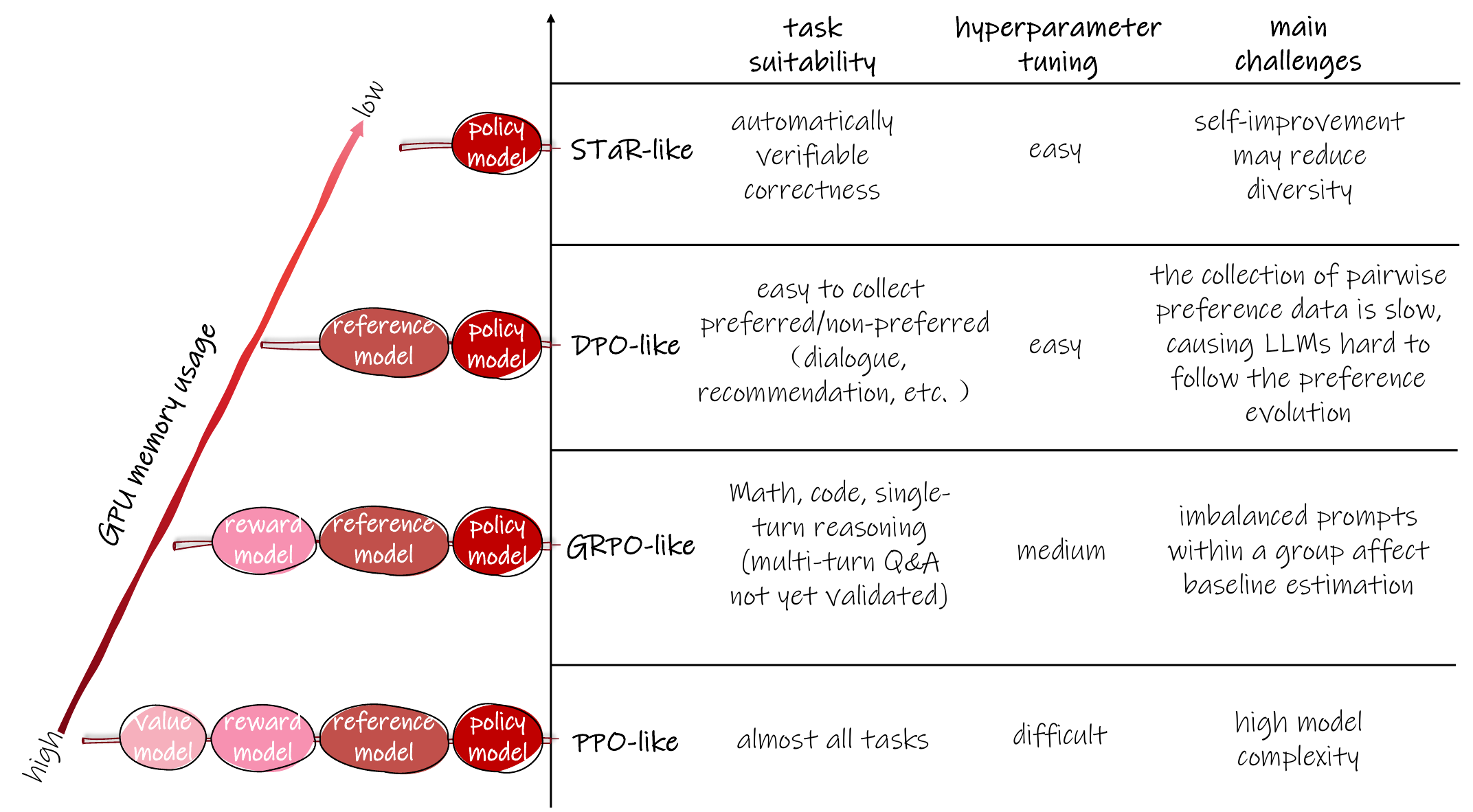}
    \caption{The comparison among RFT methods.}
    \label{fig: rft_diff}
\end{figure}

\subsubsection{Takeaways}
\begin{itemize}
    \item There are four types of RFT training methods. Their distinctions are illustrated in Figure \ref{fig: rft_diff}.
    \item Reward modeling for assessing the quality of CoT serves as the foundation of all RFT methods, typically divided into two approaches: ORM and PRM. To date, practical experience from DeepSeekR1 \cite{guo2025deepseekr1} suggests that PRM is more suitable for training smaller-scale language models, whereas ORM proves more effective for large-scale model training.
\end{itemize}

\section{Being an Insight Model}
\label{sec: insightmodel}

\begin{figure}[]
    \centering
    \includegraphics[width=0.90\linewidth]{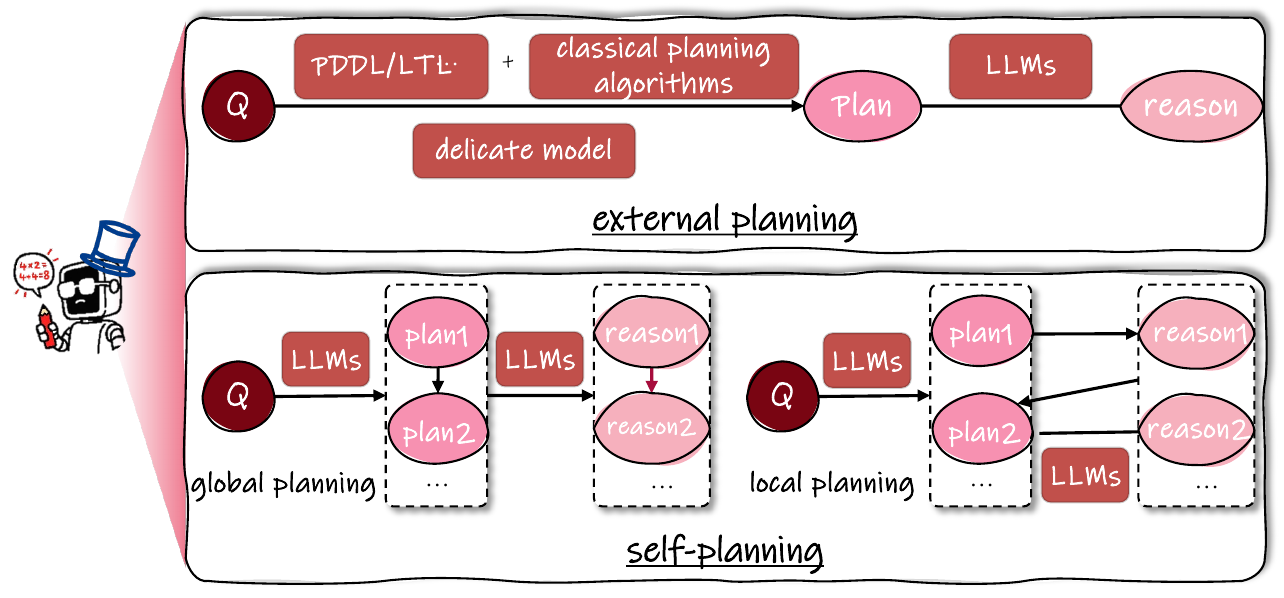}
    \caption{The framework of equipping LLMs with the planning ability.}
    \label{fig: blue}
\end{figure}

This section, from the perspective of the Six Thinking Hats, provides a detailed summary of the techniques that drive LLMs to emulate the six major human thinking abilities.

\subsection{Blue Hat} 
\label{sec: blue}
The blue hat represents the planning ability required for human reasoning. Effective planning significantly improves reasoning efficiency and quality \cite{IntelligentPlanning, wang-etal-2023-plan, norvig2002modern}. Accordingly, recent work extends the Thinking Model to strengthen LLM planning capabilities. As shown in Figure~\ref{fig: blue}, these methods fall into two categories: self-planning and external planning.

\subsubsection{Self-planning}
\label{sec: blue_1}
refers to settings where LLMs generate a reasoning plan and execute accordingly. Based on the scope of execution influenced by the plan, it can be divided into global and local planning.

Global planning formulates a complete strategy from the input task, with all subsequent steps executed strictly per the plan. CodePlan \cite{wen2024codeplan} exploits pseudocode’s expressive structure (branching, loops, modular tools, hierarchy) to generate a high-level plan before detailed reasoning, improving performance on challenging tasks. G-PlanET \cite{G-PlanET} instead uses tables as a global planning carrier. CPL \cite{wang2024cpl} further enhances global planning by using Monte Carlo Tree Search to collect step-level planning preference data.

Local planning iteratively outputs a plan, executes local reasoning, and generates subsequent plans based on intermediate results until completion. LUMOS \cite{Lumos} and Auto-GUI \cite{zhang-zhang-2024-look} collect high-quality annotations for agentic “plan-then-execute” reasoning. Jiao et al. \cite{jiao-etal-2024-learning} improve local planning via a DPO-based approach. Chern et al. \cite{chern2025Generatedthinking} extend local planning to image generation. Since plans must often satisfy logical constraints, KnowAgent \cite{zhu-etal-2025-knowagent} regulates planning steps with external knowledge bases.

A major challenge in self-planning is how to efficiently collect training data for equipping LLMs with planning abilities. To address this, methods such as AUTOACT \cite{AUTOACT}, Agentge\cite{AgentGen}, A$^{3}$T \cite{yang2024reactactre}, and STE \cite{ste} explore approaches for automatically generating planning-oriented training data.

\subsubsection{External planning}
\label{sec: blue_2} refers to leveraging external tools to perform reasoning planning. By decoupling planning from the reasoning process of LLMs, this approach circumvents the inherent limitations of LLMs in long-horizon planning. The simplest form of external planning is to rely on human-defined workflows as the planning scheme for reasoning \cite{Amor}. However, such frameworks are typically tailored to specific scenarios, lack generality, and require substantial expert effort to construct.

A more general class of methods translates natural language tasks into planning languages such as PDDL \cite{pddl} or LTL \cite{ltl}, thereby enabling the use of classical planning algorithms to generate plans. In addition, employing separately trained external models to provide planning capability while leaving reasoning to the LLM has become a common approach. For example, Cornille et al. \cite{cornille2024learning} proposed an unsupervised method to train a planner module independent of the LLM, while CoPlanner \cite{wang2024cooperativestrategicplanningenhances} applies PPO to reinforcement-tune a standalone planner.


\begin{figure}[]
    \centering
    \includegraphics[width=0.90\linewidth]{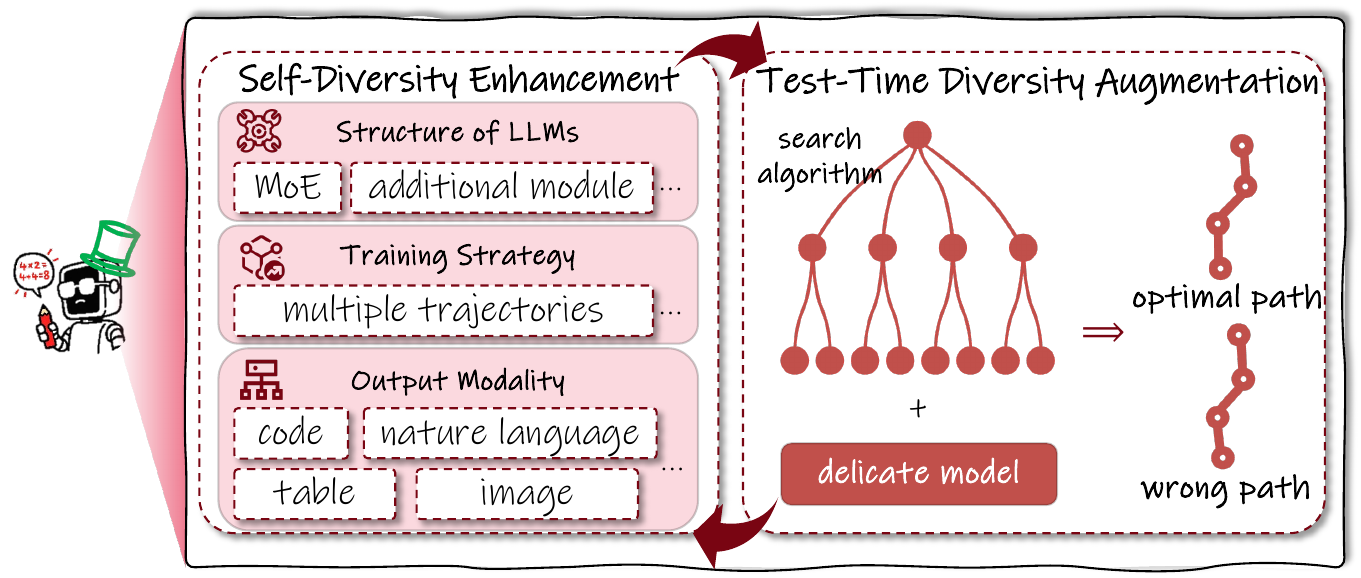}
    \caption{The framework of equipping LLMs with diverse thinking. The arrows indicate the mutually reinforcing relationship between the two approaches.}
    \label{fig: green}
\end{figure}

\subsection{Green Hat}
\label{sec: green}



The green hat symbolizes the human capacity for diverse thinking. Such diversity allows examining questions from multiple perspectives, generating varied hypotheses, and expanding the solution space \cite{DBLP:journals/corr/abs-2405-13012, liang-etal-2024-encouraging, DRDT, LLMRG}. As shown in Figure \ref{fig: green}, diverse thinking in LLMs develops along two trajectories: self-diversity enhancement and test-time diversity augmentation.

\subsubsection{Self-Diversity Enhancement} 
\label{sec: green_1}
This line of research strengthens LLMs’ inherent diverse-reasoning ability by modifying internal architectures, training strategies, and output modalities.

Architecturally, mixture-of-experts (MoE) models \cite{MoE, shen2024mixtureofexperts, MoELoRA, zhang-etal-2024-milora, wang-etal-2024-expert} allow adaptive activation of different expert parameters to reason from multiple perspectives. Beyond MoE, attention mechanisms and semantic generalization also contribute to diversity. CARE \cite{li-etal-2022-evade} adds attention regularization to mitigate monotonous generation. LADDER \cite{lidar} introduces a semantic lifting module that abstracts key terms into higher-dimensional representations, enabling escape from fixed reasoning patterns. Soft tokens \cite{butt2025softtokenshardtruths} probability-weight all candidate next-token embeddings into a mixture vector, with injected Gaussian noise jointly encouraging diverse trajectories.

From the training perspective, constructing multiple reasoning trajectories for the same question is commonly used \cite{ho-etal-2023-teacher, yuan2024advancing, chia-etal-2024-reasoning}, enabling LLMs to view questions from multiple angles. Training objectives follow SFT and RFT, either directly fine-tuning on diverse trajectories \cite{ho-etal-2023-teacher, yuan2024advancing} or using RFT feedback on alternative processes \cite{chen2025passktrainingadaptivelybalancing, walder2025passkpolicyoptimizationsolving}. ASPO \cite{ASPO} further enhances diversity via asymmetric importance sampling, preventing high-probability tokens from dominating updates while avoiding gradient collapse on low-probability tokens.

Regarding output modality, heterogeneous reasoning formats enrich diversity. Representative work \cite{Mixture-of-Thought, yu-etal-2025-chain, puerto-etal-2025-fine} trains LLMs to express reasoning in natural language, code, and tabular formats, increasing diversity of reasoning paths.

\subsubsection{Test-Time Diversity Augmentation} 
\label{sec: green_2}
At test time, varying decoding hyperparameters (e.g., temperature) can produce different reasoning outputs. Accordingly, decoding strategies such as beam search \cite{Freitag_2017}, tree search (especially Monte Carlo Tree Search \cite{browne2012survey}), and graph search \cite{corneil2008unified} can generate diverse reasoning paths for the same question. AdaReasoner \cite{AdaReasoner} further adjusts decoding hyperparameters by question type to overcome rigid reasoning patterns. To aggregate diverse paths, SSA \cite{ssa} combines sampled reasoning processes into a final answer. However, the search space induced by such decoding strategies is typically enormous, making it challenging to ensure that the sampled reasoning paths are useful for problem-solving. Thus, a key direction is using external reward models to guide exploration. Reward models, As discussed in the previous section, can assign validity scores to explored nodes, enabling pruning of invalid nodes and paths \cite{gsm8k, prm800, Math-Shepherd, ovm, li-etal-2023-making}, focusing exploration on promising trajectories.

Overall, self-diversity enhancement and test-time augmentation are mutually reinforcing: diverse LLMs yield higher-quality sampled paths, which improve the effectiveness and efficiency of test-time augmentation, while sampled test-time paths can further train LLMs to reason from multiple views.

\begin{figure}[]
    \centering
    \includegraphics[width=0.90\linewidth]{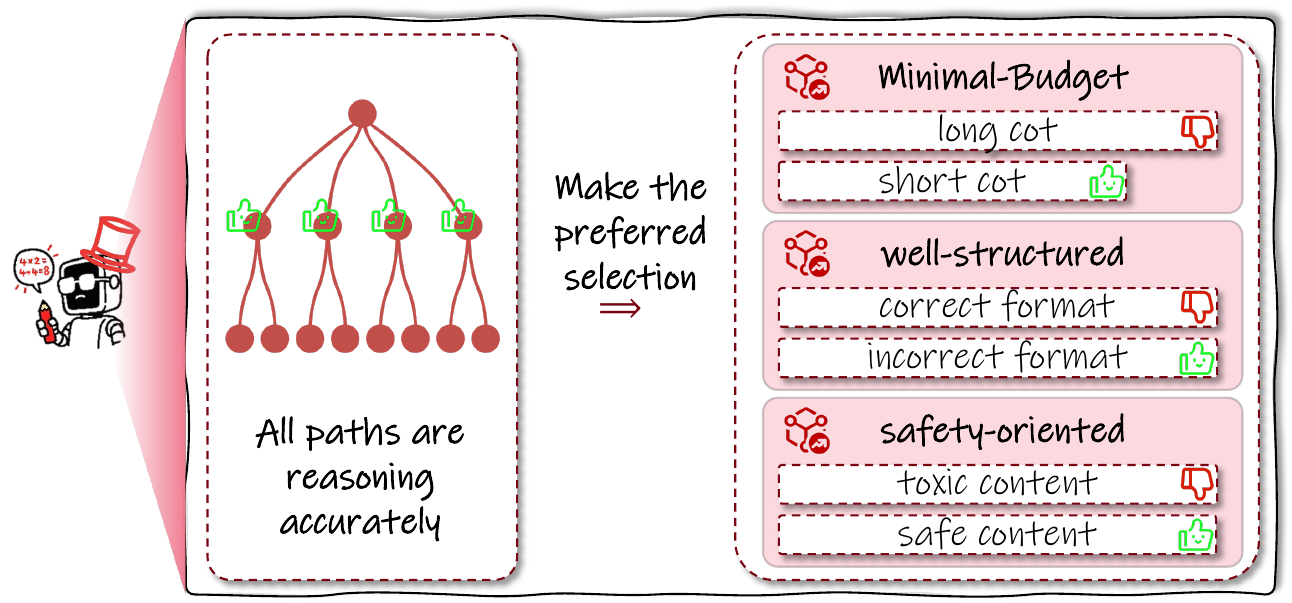}
    \caption{The illustration of how LLMs make a preferred selection.}
    \label{fig: red}
\end{figure}

\subsection{Red Hat}
\label{sec: red}





With diverse reasoning abilities, a model can generate multiple feasible solutions to a problem. However, when information is insufficient—e.g., reward models assign nearly identical scores or are unavailable in resource-constrained settings—LLMs struggle to prioritize which solution to explore. Humans, in contrast, express preferences based on knowledge and task requirements to narrow the search space and improve success under information scarcity. Inspired by this mechanism, recent studies (Figure \ref{fig: red}) investigate aligning LLMs with human-preferred reasoning patterns that aid problem solving.

\subsubsection{Minimal-budget preference}
\label{sec: red_1}
A primary preference is minimal-budget reasoning. For the same problem, LLMs may produce detailed or concise reasoning, incurring different computational costs. Minimal-budget preference guides models to select shorter reasoning paths while maintaining correctness. Park et al. \cite{park-etal-2024-disentangling} add a length penalty to the DPO objective to encourage conciseness. L1 \cite{aggarwal2025l} proposes Length-Controlled Policy Optimization to enforce user-specified length limits. Han et al. \cite{chen2025tokenbudget} introduce a token-budget-aware framework that adapts reasoning length to problem complexity, echoed by budget guidance \cite{li2025steeringllmthinkingbudget}. AutoRule \cite{AutoRule} automatically extracts human reasoning patterns to train LLMs toward clearer, more structured reasoning preferences.

\subsubsection{Well-structured reasoning formats}
\label{sec: red_2}
A second preference concerns well-structured reasoning formats. Logic-RL \cite{logicrl} and POSTERIOR-GRPO \cite{fan2025posteriorgrporewardingreasoningprocesses} add format rewards to the RFT objective, encouraging reasoning in the form $<think>...</think><answer>...</answer>$ to facilitate key information extraction. Similarly, ReSearch \cite{chen2025researchlearningreasonsearch} augments GRPO with format rewards, incentivizing LLMs to generate queries in the $<search>...</search>$ format when invoking search tools.

\subsubsection{Safety-oriented reasoning}
\label{sec: red_3}
A third preference is safety-oriented reasoning, which reduces toxic or biased outputs. Constitutional AI \cite{constitutionalai} encodes safety rules (e.g., toxicity, bias) as initial filters to align with human values. RBR \cite{mu2024rule} uses a rule-preference table with explicit symbolic constraints to balance safety and usability.

Beyond these, task-specific preferences have also been explored. For example, in tool-use scenarios, certain methods \cite{su2025pixelreasonerincentivizingpixelspace} design rewards to encourage active tool invocation rather than relying solely on internal reasoning.

\begin{figure}[]
    \centering
    \includegraphics[width=0.90\linewidth]{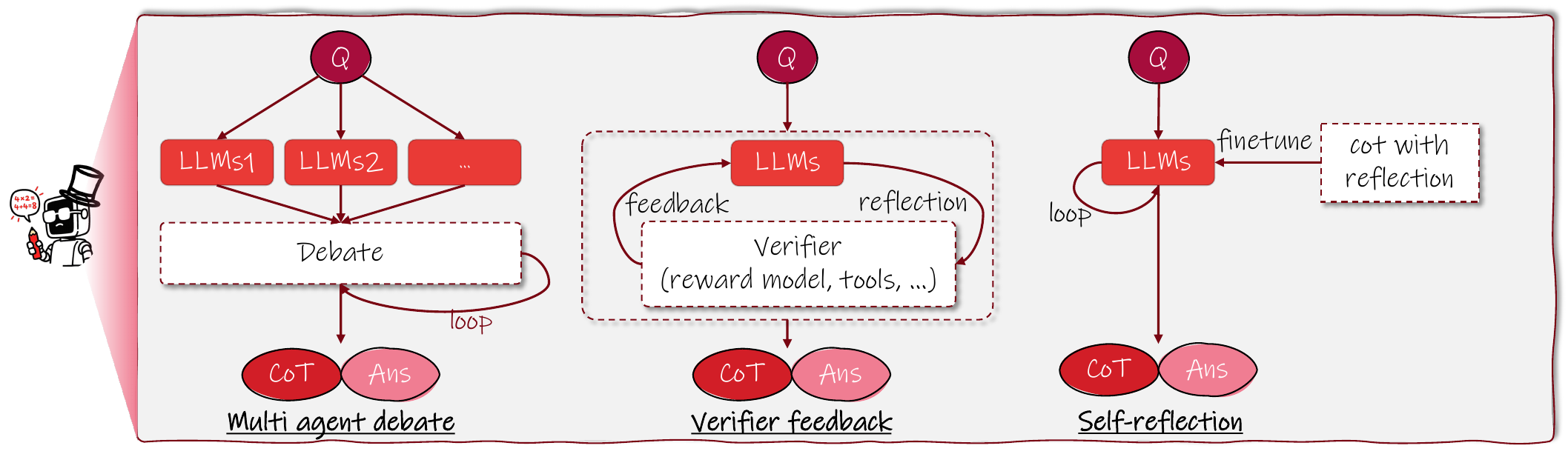}
    \caption{The main technical routes to enhance LLMs' reflective ability.}
    \label{fig: black}
\end{figure}

\subsection{Black Hat}
\label{sec: black}
The black hat represents enabling LLMs to emulate human reasoning by reviewing, questioning, debating, and refining their outputs. As shown in Figure \ref{fig: black}, current research follows three main paths: multi-agent debate, verifier feedback, and self-reflection.

\subsubsection{Multi-Agent Debate (MAD) \cite{liang-etal-2024-encouraging}} 
\label{sec: black_1} Multi-Agent Debate (MAD) involves multiple agents with distinct characteristics interacting, exchanging outputs, and providing feedback according to a communication protocol \cite{tillmann2025literaturereviewmultiagentdebate}. By leveraging diverse perspectives of different agents, MAD compensates for the cognitive limitations of a single agent. Its contributions to enhancing LLM reflection can be summarized in two aspects.
\textbf{First,} MAD facilitates the construction of training data for fine-tuning reflective abilities \cite{lan2025training, zhou-etal-2025-debate, he2025debatingtruthdebatedrivenclaim}. For example, MultiCritique \cite{lan2025training} collects CoT data including reflective processes and evaluations, then applies a two-stage training to improve reflection. D\&R \cite{zhou-etal-2025-debate} uses smaller models and LLMs in MAD to identify and correct errors, generating reflection-oriented data to enhance reasoning in smaller models.
\textbf{Second,} during inference, MAD is employed to reflect on and refine each reasoning step of LLMs \cite{estornell2025acccollab, ma2024coevolving, li-etal-2025-advancing, yang2025learningdeliberatemetapolicycollaboration, liao2025marftmultiagentreinforcementfinetuning, subramaniam2025multiagent}. Specifically, ACC-debate \cite{estornell2025acccollab} pairs an Actor (answer generator) and Critic (feedback provider) to iteratively refine reasoning. Similarly, CORY \cite{ma2024coevolving} duplicates models into generation and critique agents. MLC \cite{li-etal-2025-advancing} introduces role embeddings for autonomous role assignment. MPDF and SoftRankPO \cite{yang2025learningdeliberatemetapolicycollaboration} equip agents with metacognitive abilities to assess their own confidence, evaluate the reliability of teammates, and select the most appropriate action. MARFT \cite{liao2025marftmultiagentreinforcementfinetuning} coordinates agents via a dependency graph for efficient discussions.

\subsubsection{Verifier feedback} 
\label{sec: black_2} Verifier-feedback methods can be categorized by verifier type. The most common is reward-model-based verifiers, which not only narrow the search space but also signal whether reasoning aligns with expectations, enabling LLMs to refine their processes. Liu et al. \cite{liu2025inferencetimescalinggeneralistreward} propose self-principled critique tuning for improved inference-time scalability, while Zhong et al. \cite{zhong2025solvedetectverifyinferencetimescalingflexible} adopt a slow–fast thinking paradigm to enhance reflective efficiency.

External-tool feedback also fosters reflection. In mathematical reasoning, Leanabell-Prover \cite{zhang2025leanabellproverposttrainingscalingformal, ji2025leanabellproverv2verifierintegratedreasoningformal} leverages real-time multi-round feedback from the Lean 4 theorem prover to conduct reinforcement learning, thereby enhancing the reflective ability of LLMs in theorem proving. In medicine, HuatuoGPT-o1 \cite{chen-etal-2025-towards-medical} constructs a dataset of 40,000 verifiable medical questions as the external knowledge base to provide feedback. I For code reasoning, START \cite{li2025startselftaughtreasonertools} iteratively debugs generated code via interaction with a code interpreter.

\subsubsection{Self-reflection} 
\label{sec: black_3} This line of methods enhances the reflective ability of LLMs by constructing CoT training data that explicitly incorporates reflection processes. RISE \cite{qu2024recursive} leverages the discrepancy between reasoning conclusions and annotated answers as cues to prompt LLMs to revise their CoT, thereby generating multi-round CoT revision data. Zhang et al. \cite{zhang-etal-2024-small} and ReflectEvo \cite{li-etal-2025-reflectevo} design a prompt-based pipeline that guides LLMs to generate reflective CoT training data. Yang et al. \cite{yang2025supercorrect} and PTR \cite{du2025think} collect CoT with reflection processes by leveraging collaboration between larger and smaller models. Welleck et al. \cite{welleck2023generating} and SPOC \cite{zhao2025boosting} decompose the reasoning process into an iterative think-reflection procedure to gather data.

\begin{figure}[]
    \centering
    \includegraphics[width=0.90\linewidth]{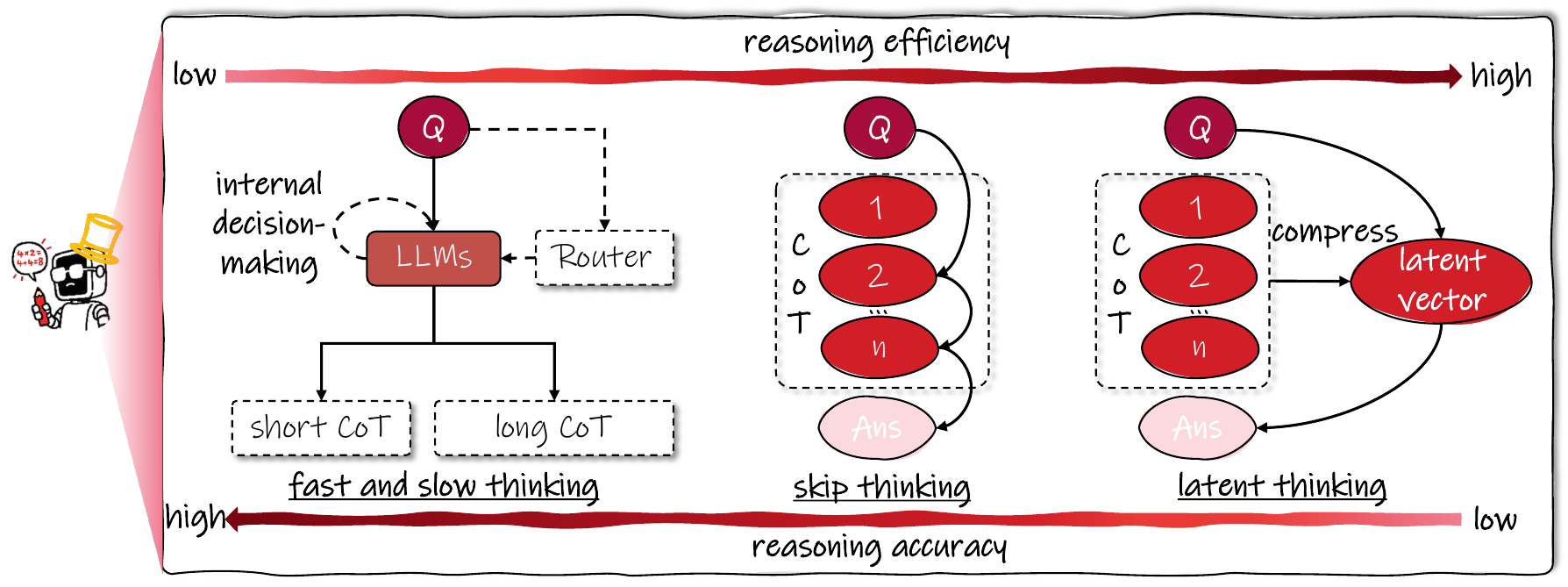}
    \caption{Representative techniques for speeding up cot reasoning.}
    \label{fig: yellow}
\end{figure}

\subsection{Yellow Hat}
\label{sec: yellow}
The yellow cap symbolizes the optimistic facet of human reasoning. Excessive reflection can hinder efficiency; thus, when facing high-certainty problems, humans tend to streamline their reasoning for faster outcomes. Inspired by this, many studies have sought to accelerate reasoning, which can be categorized into three sub-directions (Figure \ref{fig: yellow}): integration of slow and fast thinking, skip-thinking, and latent thinking.

\subsubsection{Integration of slow and fast thinking}
\label{sec: yellow_1}
Fast thinking corresponds to concise reasoning processes, while slow thinking corresponds to more elaborate ones \cite{kahneman2011thinking, khalil2025adaptive}. LLMs generally possess both modes. For example, with a prompt designed to generate a draft chain \cite{xu2025chaindraftthinkingfaster}, an LLM may abandon lengthy chains of thought and instead reason using concise expressions. However, LLMs cannot autonomously determine under which circumstances each mode should be applied, and thus typically default to slow thinking to ensure accuracy. To accelerate reasoning, existing approaches investigate two directions for enabling LLMs to adaptively select the appropriate reasoning mode. The first category involves \textbf{internal decision-making} within the model \cite{yu2025longshortchainofthoughtmixturesupervised, luo2025autol2sautolongshortreasoning, zhang2025synapserouteautorouteswitchingframework, zhao2025letllmsbreakfree, zhang2025adaptthinkreasoningmodelslearn, park-etal-2024-disentangling, aggarwal2025l, chen2025tokenbudget, jiang2025think}. For instance, Yu et al. \cite{yu2025longshortchainofthoughtmixturesupervised} train LLMs with a mixture of short and long reasoning chains to develop adaptive mode-selection capabilities, while Luo et al. \cite{luo2025autol2sautolongshortreasoning} extend this approach by introducing indicator tokens to make the decision process explicit. Beyond SFT, RFT with reward mechanisms that balance answer accuracy and reasoning conciseness \cite{park-etal-2024-disentangling, aggarwal2025l, chen2025tokenbudget, zhang2025adaptthinkreasoningmodelslearn} can also encourage LLMs to choose appropriate reasoning modes based on the task. The second category relies on \textbf{external modules for guidance}. External modules can be divided into hand-crafted heuristic rules \cite{yan2025murmomentumuncertaintyguided} and trainable router modules \cite{11127195, liang2025thinkswitcherthinkhardthink, yao2025hdflow, he2025selfrouteautomaticmodeswitching}. Since the former one does not involve fine-tuning, this survey focuses on the latter. In this direction, ThinkSwitcher \cite{liang2025thinkswitcherthinkhardthink} uses query embeddings to predict the success rates of different reasoning modes and select accordingly, while HDFlow \cite{yao2025hdflow} employs a hybrid approach: beginning with fast thinking and switching to slow thinking if the verifier detects an error. Self-Route \cite{he2025selfrouteautomaticmodeswitching} and Adaptive Thinking \cite{11127195} train classifiers to select between slow and fast reasoning modes based on problem complexity.

Fast thinking corresponds to concise reasoning processes, whereas slow thinking involves more elaborate ones \cite{kahneman2011thinking, khalil2025adaptive}. LLMs generally support both modes. For example, a draft-chain prompt \cite{xu2025chaindraftthinkingfaster} can encourage concise reasoning for LLMs. However, LLMs cannot autonomously determine which mode to apply and often default to slow thinking to ensure accuracy. To accelerate reasoning, methods focus on two directions for adaptive mode selection. The first is internal decision-making, where LLMs learn to choose modes autonomously \cite{yu2025longshortchainofthoughtmixturesupervised, luo2025autol2sautolongshortreasoning, zhang2025synapserouteautorouteswitchingframework, zhao2025letllmsbreakfree, zhang2025adaptthinkreasoningmodelslearn, park-etal-2024-disentangling, aggarwal2025l, chen2025tokenbudget, jiang2025think}. For instance, Yu et al. \cite{yu2025longshortchainofthoughtmixturesupervised} train LLMs with a mixture of short and long reasoning chains to develop adaptive mode-selection capabilities, while Luo et al. \cite{luo2025autol2sautolongshortreasoning} extend this approach by introducing indicator tokens. Beyond SFT, RFT with reward mechanisms that balance answer accuracy and reasoning conciseness \cite{park-etal-2024-disentangling, aggarwal2025l, chen2025tokenbudget, zhang2025adaptthinkreasoningmodelslearn} can also encourage LLMs to choose appropriate reasoning modes based on the task. The second category relies on \textbf{external modules for guidance}. External modules can be divided into hand-crafted heuristic rules \cite{yan2025murmomentumuncertaintyguided} and trainable router modules \cite{11127195, liang2025thinkswitcherthinkhardthink, yao2025hdflow, he2025selfrouteautomaticmodeswitching}. Since the former one does not involve fine-tuning, this survey focuses on the latter. In this direction, ThinkSwitcher \cite{liang2025thinkswitcherthinkhardthink} uses query embeddings to predict the success rates of different reasoning modes and select accordingly, while HDFlow \cite{yao2025hdflow} employs a hybrid approach: beginning with fast thinking and switching to slow thinking if the verifier detects an error. Self-Route \cite{he2025selfrouteautomaticmodeswitching} and Adaptive Thinking \cite{11127195} train classifiers to select between slow and fast reasoning modes based on problem complexity.

\subsubsection{Skip-thinking}
\label{sec: yellow_2} Skip-thinking accelerates LLM inference by bypassing less critical parts of the reasoning process. Skip-thinking \cite{chen2025skip} leverages answer feedback to train LLMs to skip unimportant reasoning chunks. Subsequent work refines this granularity: Liu et al. \cite{liu2024can} target reasoning steps, while Xia et al. \cite{xia2025tokenskipcontrollablechainofthoughtcompression} focus on individual output tokens.

\subsubsection{Latent thinking}
\label{sec: yellow_3} Recent work has explored training LLMs to reason in latent space \cite{deng2023implicitchainthoughtreasoning, goyal2024thinkspeaktraininglanguage, deng2025from, hao2024traininglargelanguagemodels, su2025token, chen2026ImgCoT}. The core idea is to encode explicit rationales into latent representations, enabling implicit reasoning and direct answer generation. For instance, Deng et al. \cite{deng2025from} progressively remove reasoning traces during training, embedding rationales into the latent space, while Chen et al. \cite{chen2026ImgCoT} compress CoT into a compact visual latent space. However, these methods may achieve lower accuracy than explicit reasoning approaches in most cases.

\begin{figure}[]
    \centering
    \includegraphics[width=0.90\linewidth]{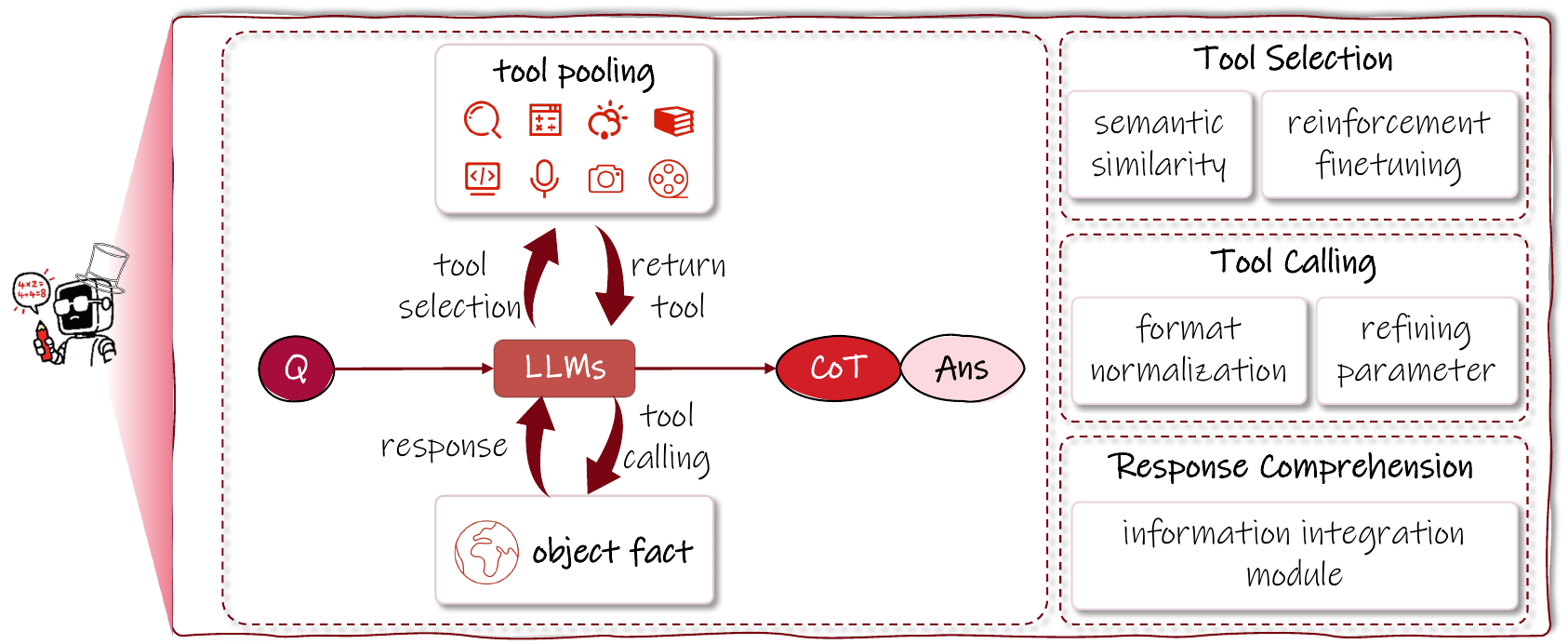}
    \caption{The pipeline of adopting tools to perceive objective facts.}
    \label{fig: white}
\end{figure}

\subsection{White Hat}
\label{sec: white}
The white hat denotes the capacity to perceive objective facts. In LLMs, this ability is advanced through the use of external tools \cite{qu2025tool}, which extend knowledge boundaries (e.g., search engines, knowledge bases), enhance domain-specific skills (e.g., Python interpreters), and improve interactive capabilities (e.g., speech recognition, image analysis), enabling perception of the physical world. The development of tool usage can be categorized into three directions corresponding to different stages of tool integration (Figure \ref{fig: white}).



\subsubsection{Tool Selection}
\label{sec: white_1}
A simple and effective strategy for tool selection is to compute semantic similarity between task and tool descriptions \cite{reimers-gurevych-2019-sentence, lei-etal-2023-unsupervised,10.1561/1500000019, zheng-etal-2024-toolrerank, 10.1145/3627673.3679847}, ranking tools to select the most suitable one. However, this approach typically selects only a single tool, limiting compositionality. To enable tool combination, Vista \cite{huang2025visualtoolagent} decouples tool selection into a separate small model trained with answer correctness feedback, thereby improving multi-tool coordination.

\subsubsection{Tool Calling}
\label{sec: white_2}Once appropriate tools are identified, the next challenge is effective tool calling. Existing approaches include constructing instruction datasets to fine-tune LLMs for standardized tool usage \cite{yang2023gpt4tools}, introducing special tokens to disentangle parameter generation from reasoning \cite{NEURIPS2023_8fd1a81c}, and applying RFT to improve parameter accuracy \cite{li2024toolaugmented}. Beyond enhancing autonomous invocation, some works propose standardized protocols to enable seamless interaction with tools featuring heterogeneous interfaces \cite{openai2023functioncalling, langchain2022framework, liu2022llamaindex, anthropic2024mcpintro}.

\subsubsection{Response Comprehension}
\label{sec: white_3} After tool calling, interpreting complex feedback (such as tables, code, or images) and expressing it in natural language is essential for perceiving objective facts. ReCOMP \cite{xu2024recomp} compresses verbose information into concise, relevant content, while ConAgents \cite{shi-etal-2024-learning} trains an observation agent to dynamically generate functions for targeted information extraction.


\begin{table*}
    \centering
    \scriptsize
    \setlength{\tabcolsep}{2.4mm}
    \caption{Typical benchmark datasets for LLMs reasoning.}
    \label{tab: benchmarks}
    \begin{tabular}{ccccccc} \hline
        Category    & Dataset  & Train & Valid & Test  & Task Description &  CoT  \\ \hline
        \multirow{4}{*}{General Task}    & BBH \cite{BBH}  & 0 & 0&6511 & 23 common reasoning tasks  & No\\
                                    & SuperGPQA \cite{SuperGPQA}  & 0 &  0 &  26,529 & Graduate-level Q\text{\&}A tasks across 285 disciplines  & No \\
                                    & MMLU\cite{hendrycks2021measuring} & 0&1,540 &14368 & Multiple-choice Q\&A across 57 tasks & No \\
                                    & MMLUPro \cite{mmlupro}  & 0 &  70 & 12032 & Q\text{\&}A tasks across 285 disciplines & Partial \\ \hline
        \multirow{6}{*}{Mathematics} & GSM8K \cite{gsm8k} & 7500 &  0 &1000 & Grade school math & Yes \\
                                     & MGSM \cite{mgsm} & 88 &  0 &2750 & Multilingual version of GSM8K & Yes \\
                                     & AQuA \cite{ling-etal-2017-program} & 100,949 & 250 & 250 & Algebraic word problems & Yes \\
                                     & MATH \cite{math}  & 7500 &  0 &5000 & Competitive math & Yes \\ 
                                     & AIME \cite{AIME} & 0 & 0 & Updated annually & American invitational mathematics examination & Yes \\
                                     & Geometry3K \cite{lu-etal-2021-inter} & 2101& 300&601 & Geometry Problem with symbolic reasoning & No \\
                                     \hline
        \multirow{7}{*}{Coding}   & CodeContests \cite{codecontests} & 13328 &  117 &165 & Competitive programming & No \\ 
                                  & LiveCodeBench \cite{livecodebench} & 0   & 0  &   Continuous updates   &  Periodically updated programming & No \\
                                  & MHPP \cite{dai2025mhpp} &  0  &  0 &   210  & Manually created Python programming   & No  \\
                                  & EquiBench \cite{EquiBench} &  0  & 0  &  2400   &  Equivalence checking for two programs  & No \\
                                  & MBPP Pro \cite{MBPPPro} &  0  & 0  &   378  &  Self-invoking code generation  & No \\
                                  & SWEbench \cite{jimenez2024swebench} &  19,000  & 0  &  2294   & Solving real-world GitHub issues  & No \\ 
                                  & BFCL v3 \cite{BFCL} &  0  & 0  &  2000   & Function-calling tasks  & No \\
                                  \hline
        \multirow{4}{*}{Commonsense}  & HotpotQA \cite{yang-etal-2018-hotpotqa} & 90564 &  7405 & 14810 & Reading comprehension & No \\
                                    & CommonsenseQA \cite{commonsenseqa} & 9797 &  1225 & 1225 & Multiple-choice  Q\text{\&}A & No \\
                                    & StrategyQA \cite{StrategyQA} & 2290 &  0 & 490 & True or false  Q\text{\&}A & Yes \\
                                    & OpenBookQA \cite{OpenBookQA} & 13328 &  117 &165 & Multiple-choice  Q\text{\&}A & No \\ \hline
        
        \multirow{4}{*}{Domain Knowledge}    & MedQA \cite{medQA} & 48876 & 6109&6112 & Multilingual multiple-choice medical Q\text{\&}A & No\\
                                    & JAMA \cite{JAMA} & 0 &  0 &  1524 & Multiple-choice clinical  Q\text{\&}A  & Yes \\
                                    & MedXpertQA \cite{zuo2025medxpertqa} & 0 &  0 & 4460 & Multimodal multiple-choice medical Q\text{\&}A & Yes \\
                                    & GPQA \cite{rein2024gpqa} & 0 & 0& 448& Biology, physics, and chemistry multiple-choice Q\text{\&}A   & Yes \\ \hline
        \multirow{6}{*}{Others}    & ZebraLogic \cite{lin2025zebralogic} & 0 & 0&1000 & Logic grid puzzles & No\\
                                    & ToolBench \cite{toolllm} & 0 &  0 &  1524 & General tool-use tasks  & Yes \\
                                    & ALFWorld \cite{shridhar2021alfworld} & 3553 &0 & 274 & 6 types of decison making tasks  & No \\ 
                                    & ChartQA-H\cite{masry-etal-2022-chartqa} &  7398 & 960 & 1250 & Charts with visual and logical reasoning & No \\
                                    & ChartQA-M\cite{masry-etal-2022-chartqa}  & 20901 & 960 & 1250 & Charts with visual and logical reasoning & No \\
                                    \hline
    \end{tabular}
\end{table*}

\subsection{Takeaways}
%
\begin{itemize}
    \item Different hats correspond to distinct human reasoning mechanisms. By emulating these mechanisms, LLMs progressively approximate human-like thinking. Specific implementations are illustrated in Figures \ref{fig: blue} - \ref{fig: white}.
    \item The techniques for constructing these six reasoning mechanisms still have limitations, which will be discussed in detail in Section \ref{sec: chanllenge}.
\end{itemize}

\section{Evaluations of CoT Finetuning}
\label{sec: datasets}
This section first offers a comprehensive summary of datasets (available in our GitHub repository and updated periodically) for evaluating the reasoning capabilities of LLMs.

\subsection{Benchmarks}
Benchmarks for evaluating large language models can be categorized into six types according to their primary focus on reasoning capabilities: mathematics, code, commonsense, domain knowledge, and others. Representative benchmarks for each category are summarized in Table \ref{tab: benchmarks}. These benchmarks are often used in combination to comprehensively assess the reasoning abilities of LLMs across multiple dimensions.

\subsection{Performance of representative methods}
We present the performance of the representative methods corresponding to the six thinking-hat capabilities on common tasks (blue hat – Figure \ref{tab: acc_blue}, green hat – Figure \ref{tab: acc_green}, red hat – Figure \ref{tab: acc_red}, black hat – Figure \ref{tab: acc_black}, yellow hat – Figure \ref{tab: acc_yellow}), to facilitate researchers in: 1) understanding the commonly adopted base LLMs, and 2) selecting appropriate baselines based on performance. Note, for the white hat, we do not provide performance comparison since most methods have not been evaluated under unified tasks. ``-" in the figures means that the result is not reported by the corresponding methods.

\section{CHALLENGES \& FUTURE DIRECTIONS}
\label{sec: chanllenge}


\begin{table}[]
\centering
\scriptsize
\setlength{\tabcolsep}{0.55mm}
\caption{Accuracy of different methods with \textcolor{blue}{planning ability}.}
\begin{tabular}{ccccc}
\hline
Methods              & Base Model               & GSM8K & HotpotQA & ALFWorld \\ \hline
\multirow{3}{*}{Codeplan}\cite{wen2024codeplan}  & Mistral-7B               & 59.5  & 40.4     & 23.2     \\
                           & Llama-2-7B               & 33.8  & 27.4     & 14.1     \\
                           & Llama-2-13B              & 49.5  & 40.4     & 33.3     \\ \hline
\multirow{2}{*}{pDPO}\cite{jiao-etal-2024-learning}      & Gemma-2B                 & 52.8  & -        & -        \\
                           & DeepSeekMath-7B          & 82.3  & -        & -        \\ \hline
\multirow{3}{*}{KnowAgent} \cite{zhu-etal-2025-knowagent} & Llama-2-7B               & -     & 33.5     & 29.3     \\
                           & Llama-2-13B              & -     & 39.3     & 54.3     \\
                           & Llama-2-70B              & -     & 48.1     & 77.1     \\
\multirow{3}{*}{AUTOACT \cite{AUTOACT}}   & Mistral-7B-Instruct-v0.2 & -     & 38.9     & -        \\
                           & Llama-2-13B              & -     & 40.5     & -        \\
                           & Llama-2-70B              & -     & 48.5     & -        \\ \hline
CPL\cite{wang2024cpl}                        & DeepSeekMathBase-7B      & 73.8  & -        & -        \\ \hline
\multirow{2}{*}{AGENTGEN \cite{AgentGen}}  & LLAMA-3.1-8B             & -     & -        & 17.9     \\
                           & LLAMA-3.1-70B            & -     & -        & 19.4     \\ \hline
\multirow{2}{*}{AMOR \cite{Amor}}      & Llama-2-7B               & -     & 45.8     & -        \\
                           & Llama-2-13B              & -     & 48.6     & -        \\ \hline
\multirow{2}{*}{LUMO \cite{Lumos}}     & LLAMA-2-7B               & 50.5  & -        & 23.5     \\
                           & LLAMA-2-13B              & 55.4  & -        & 50.2     \\ \hline
\end{tabular}
\label{tab: acc_blue}
\end{table}

\subsection{Blue cap: Meta planning}
Existing studies focus on “how to plan for a task” rather than “planning itself,” lacking an abstract, general meta-planning capability \cite{stefik1981planning, wilensky1981meta}. Thus, planning remains at a task-oriented level rather than a cognition-oriented strategy level.

Advancing meta-planning presents a key research opportunity. If LLMs can acquire task-independent planning strategies—such as selecting appropriate decomposition methods in unfamiliar environments, dynamically adjusting plans, and leveraging feedback to correct trajectories—they may significantly improve cross-task reasoning and adaptability \cite{xiong2025mpoboostingllmagents}. This would enhance robustness and interpretability in open environments and may represent an important step toward human-like cognition.

\begin{table}[]
\centering
\scriptsize
\setlength{\tabcolsep}{0.55mm}
\caption{Accuracy of different methods with \textcolor{green}{diverse thinking} .}
\begin{tabular}{cccc}
\hline
Methods                        & Base Model           & GSM8K & MATH \\ \hline
\multirow{2}{*}{EURUS \cite{yuan2024advancing}}         & Mistral-7B           & -     & 34.2 \\
                               & CodeLLaMA-70B        & -     & 41.7 \\ \hline
RPO \cite{chia-etal-2024-reasoning}                            & LLaMA-3-8B           & 64.2  & 22.2 \\ \hline
CoR  \cite{yu-etal-2025-chain}                          & DeepSeekMath-Base7B  & 88.7  & 66.7 \\ \hline
\multirow{3}{*}{DCoT \cite{puerto-etal-2025-fine}}          & Llama-2-7b           & 29.6  & -    \\
                               & Llama-2-13b          & 44.3  & -    \\
                               & Llama-2-70b          & 66.0  & -    \\ \hline
\multirow{3}{*}{Soft tokens \cite{butt2025softtokenshardtruths}}          & Llama-3.2-3B           & 77.2  & -    \\
                               & Llama-3.1-8B          & 82.6  & -    \\
                               & Qwen-2.5-3B          & 82.9  & -    \\ \hline
\multirow{2}{*}{PKPO\cite{walder2025passkpolicyoptimizationsolving}}          & GEMMA2-9B            & -     & 48.4 \\
                               & LLAMA3.1-8B          & -     & 61.9 \\ \hline
\multirow{3}{*}{SSA \cite{ssa}}           & Qwen-2.5-base-0.5B   & 92.7  & 75.4 \\
                               & Qwen-2.5-base-1B     & 92.5  & 76.6 \\
                               & Qwen-2.5-base-3B     & 93.3  & 76.8 \\ \hline
\multirow{2}{*}{Math-shepherd \cite{Math-Shepherd}} & Mistral-7B           & 84.1  & 33.0 \\
                               & LLaMA2-7B            & 73.2  & 21.6 \\ \hline
\multirow{2}{*}{OVM \cite{ovm}}           & Mistral-7B           & 84.7  & -    \\
                               & Llama2-7B            & 73.7  & -    \\ \hline
\multirow{2}{*}{DIVERSE \cite{li-etal-2023-making}}       & GPT-3 davinci (175B) & 30.9  & -    \\
                               & code-davinci-002     & 82.3  & -    \\ \hline
\end{tabular}
\label{tab: acc_green}
\end{table}

\subsection{Green cap: From surface to internal}
Current work on reasoning diversity in LLMs primarily adjusts decoding hyperparameters (e.g., temperature) to produce variation in surface-level expressions. For example, when answering “Why does it rain?”, the LLMs may generate outputs such as “Because water vapor in the air condenses into raindrops” or “Rain forms when water vapor cools and falls from clouds,” yet these represent only surface linguistic diversity. In contrast, true diversity should reflect methodological differences, such as employing deductive, inductive, or analogical reasoning \cite{cheng2025inductive, cai-etal-2025-role}. However, models currently struggle to generate fundamentally distinct reasoning approaches, limiting flexibility and creativity.

Enhancing the intrinsic method diversity of reasoning offers substantial potential. If LLMs can autonomously adopts diverse reasoning methods for the same problem, they could better explore solution spaces in complex or information-scarce settings and advance toward human-like diverse thinking.

\subsection{Red cap: robust preference}
Incorporating preference into reasoning enables LLMs to follow preferred reasoning paths, but robustness remains challenging. A major issue is cross-task generalization \cite{li2025omnithinkerscalingcrossdomaingeneralization}: preferences such as “minimum-cost” reasoning may suit general Q\&A tasks, yet high-risk domains (e.g., law, medicine) require lengthy, repeatedly verified reasoning. Current preference reasoning learning does not sufficiently capture such domain differences. Another concern is adversarial manipulation, where preference signals can be exploited (e.g., reward hacking \cite{shihab2025detectingmitigatingrewardhacking}) to produce outputs that appear aligned but are harmful or goal-divergent.

If these robustness issues can be mitigated \cite{li2025omnithinkerscalingcrossdomaingeneralization, liu2025rrm, miao2024inform}, preference reasoning holds broad potential, such as supporting personalized cognitive pathways in education and adapting to users’ reasoning habits in human–AI collaboration. Thus, robust preference reasoning is crucial for practical deployment and for advancing safer, more personalized, and trustworthy intelligence.

\begin{table}[]
\centering
\scriptsize
\setlength{\tabcolsep}{0.45mm}
\caption{Accuracy of different methods with \textcolor{red}{intuitive judgement} . }
\begin{tabular}{ccccc}
\hline
Methods                          & Base Model                    & GSM8K & AMC  & AIME \\ \hline
L1 \cite{aggarwal2025l}                             & \makecell[c]{DeepSeek-R1-Distill-Qwen-1.5B} & -     & 46.7 & 12.1 \\ \hline
Token-Budget   \cite{chen2025tokenbudget}                  & Llama-3.1-8B                  & 29.3  & -    & -    \\ \hline
\multirow{3}{*}{Budget Guidance \cite{li2025steeringllmthinkingbudget}} & \makecell[c]{DeepSeek-R1-Distill-Qwen-7B}   & -     & 60.2 & 33.3 \\
                                 & \makecell[c]{DeepSeek-R1-Distill-Qwen-32B}  & -     & 69.9 & 56.7 \\
                                 & Qwen3-8B                      & -     & 80.7 & 50.0 \\ \hline
\end{tabular}
\label{tab: acc_red}
\end{table}

\subsection{Black cap: stop overthinking}
Incorporating reflection into LLM reasoning aims to improve monitoring and error correction. However, current models often perform exhaustive reflection even on simple or deterministic tasks, leading to over-reflection \cite{sui2025stop}, longer reasoning paths, unnecessary complexity, and increased computational cost. A key challenge is designing mechanisms that distinguish necessary from redundant reflection, triggering self-checks only when needed and producing direct outputs for low-risk steps. Such adaptive reflection could improve efficiency and reliability in complex tasks, reducing latency while maintaining reasoning quality.

\begin{table}[]
\centering
\scriptsize
\setlength{\tabcolsep}{0.55mm}
\caption{Accuracy of different methods with \textcolor{black}{timely reflection} .}
\begin{tabular}{ccccc}
\hline
Methods                             & Base Model               & GSM8K & MATH  & MMLU \\ \hline
D\&R  \cite{zhou-etal-2025-debate}                              & Mistral-7B               & -     & 17.3  & 67.2 \\ \hline
\multirow{3}{*}{ACC-COLLAB \cite{estornell2025acccollab}}         & Mistral-7B               & -     & -     & 67.2 \\
                                    & Llama-3-8B               & -     & -     & 68.3 \\
                                    & Gemma-2-2B               & -     & -     & 55.5 \\ \hline
\multirow{4}{*}{MLC \cite{li-etal-2025-advancing}}                & LLAMA-2-7B               & 33.0  & 5.0   & -    \\
                                    & LLAMA-2-13B              & 32.0  & 12.0  & -    \\
                                    & Llama-3.1-8B             & 83.6  & 54.0  & -    \\
                                    & Mistral-7B-Instruct-v0.2 & 69.0  & 16.0  & -    \\ \hline
SoftRankPO \cite{yang2025learningdeliberatemetapolicycollaboration} & Llama-3.1-8B             & 86.35 & 82.72 & 70.9 \\ \hline
MARFT \cite{liao2025marftmultiagentreinforcementfinetuning}                               & Qwen2.5-Coder-3B         & 78.7  & -     & -    \\ \hline
\multirow{2}{*}{Subramaniam et al. \cite{subramaniam2025multiagent}} & LLaMA-3-8B               & 88.6  & 57.4  & -    \\
                                    & Mistral-7B               & 58.4  & 22.5  & -    \\ \hline
START   \cite{li2025startselftaughtreasonertools}                            & QwQ-32B                  & -     & -     & -    \\ \hline
RISE \cite{qu2024recursive}                               & Mistral-7B               & 59.2  & 18.4  & -    \\ \hline
\multirow{4}{*}{ReflectEvo \cite{li-etal-2025-reflectevo}}         & Llama-3-8B               & -     & 14.4  & -    \\
                                    & Llama-3-70B              & -     & 40.8  & -    \\
                                    & Mistral-7B-Instruct-v0.2 & -     & 7.6   & -    \\
                                    & Mistral-22B              & -     & 47.4  & -    \\ \hline
\multirow{2}{*}{PTR \cite{du2025think}}                & Qwen2-7B                 & 79.9  & 48.9  & 63.2 \\
                                    & Llama3-8B                & 79.6  & 24.9  & 68.6 \\ \hline
SELF-CORRECT \cite{welleck2023generating}                       & GPT-NEO-1.3B             & 24.2  & -     & -    \\ \hline
\end{tabular}
\label{tab: acc_black}
\end{table}

\subsection{Yellow cap: trade-off between efficiency and performance}
In accelerating CoT reasoning, the main challenge is improving efficiency while maintaining accuracy and interpretability. Fast–slow thinking dynamically selects short or long chains, but misrouting may harm performance. Skip-thinking omits intermediate steps, potentially affecting logical consistency, while latent thinking compresses reasoning into compact representations, risking semantic loss and uncontrollable processes. Balancing these trade-offs can expand LLM applicability: in real-time interactive scenarios (e.g., dialogue systems, intelligent assistants) \cite{Li_2025_CVPR}, fast reasoning reduces latency and enhances user experience, and in resource-constrained settings (e.g., mobile or edge devices), efficient reasoning lowers computation and energy consumption \cite{chen2025skip}, facilitating deployment of complex models.

\begin{table}[]
\centering
\scriptsize
\setlength{\tabcolsep}{0.4mm}
\caption{Accuracy of different methods with \textcolor{orange}{internal thinking} .}
\begin{tabular}{cccccc}
\hline
\multirow{2}{*}{Methods} & \multirow{2}{*}{Base Model}   & \multicolumn{2}{c}{GSM8K} & \multicolumn{2}{c}{AIME} \\ \cline{3-6} 
                                &                               & ACC    & \makecell[c]{$\#$Token}  & ACC    & \makecell[c]{$\#$Token}  \\ \hline
LS-Mixture\cite{yu2025longshortchainofthoughtmixturesupervised}              & s1.1-32B                      & -      & -                & 60.0   & 40251           \\ \hline
\multirow{2}{*}{AutoL2S \cite{luo2025autol2sautolongshortreasoning}}     & Qwen2.5-3B                    & 82.6   & 741              & -      & -               \\
                                & Qwen2.5-7B                    & 92.9   & 488              & -      & -               \\ \hline
\multirow{4}{*}{SBT \cite{zhao2025letllmsbreakfree}}         & Qwen2.5-Math-1.5B             & 84.9   & 414              & 14.2   & 997             \\
                                & Qwen2.5-Math-7B               & 95.4   & 956              & 38.4   & 9778            \\
                                & Llama-3.2-1B                  & 41.2   & 698              & 1.0    & 6821            \\
                                & Llama-3.1-8B                  & 88.3   & 997              & 7.7    & 5845            \\ \hline
\multirow{2}{*}{LHRMs \cite{jiang2025think}}     & Qwen2.5-Math-1.5B                    & -   & -              & 35.3      & -               \\
                                & Qwen2.5-Math-7B                    & -   & -              & 66.7      & -               \\ \hline
\multirow{2}{*}{AdaptThink \cite{zhang2025adaptthinkreasoningmodelslearn}}  & \makecell[c]{DeepSeek-R1-Distill-Qwen-1.5B} & 83.1   & 480              & 31.0   & 6679            \\
                                & \makecell[c]{DeepSeek-R1-Distill-Qwen-7B}   & 91.0   & 309              & 55.6   & 8599            \\ \hline
\multirow{3}{*}{ThinkSwitcher \cite{liang2025thinkswitcherthinkhardthink} }  & \makecell[c]{DeepSeek-R1-Distill-Qwen-1.5B} & 84.7   & 2114             & 23.3   & 8192            \\
                                & \makecell[c]{DeepSeek-R1-Distill-Qwen-7B}   & 92.5   & 1389             & 48.3   & 7936            \\
                                & \makecell[c]{DeepSeek-R1-Distill-Qwen-14B}  & 94.3   & 1042             & 60.4   & 8044            \\ \hline
\multirow{3}{*}{Self-Route \cite{he2025selfrouteautomaticmodeswitching}}     & Qwen2.5-7B                    & 92.6   & 319              & 53.3   & 11357           \\
                                & Qwen2.5-32B                   & 96.4   & 317              & 73.3   & 8523.6          \\
                                & Qwen3-8B                      & 94.3   & 333              & 76.7   & 13858           \\ \hline
Skipthinking \cite{chen2025skip}                    & Llama3.2-3B                   & 80.2   & -                & -      & -               \\ \hline
TokenSkip \cite{xia2025tokenskipcontrollablechainofthoughtcompression}                       & LLaMA-3.1-8B                  & 86.7   & 214              & -      & -               \\ \hline
Coconut  \cite{hao2024traininglargelanguagemodels}                        & GPT-2-base                    & 34.1   & 8                & -      & -               \\ \hline
Token Assorted  \cite{su2025token}                        & LLama3.2-3B                    & 73.8   & 198.8                & -      & -               \\ \hline
ImgCoT  \cite{hao2024traininglargelanguagemodels}                        & LLama3.2-3B                    & 56.8   & 8                & -      & -               \\ \hline
\end{tabular}
\label{tab: acc_yellow}
\end{table}

\subsection{White cap: perception of complete objective information}
Although LLMs have advanced in using tools to perceive objective facts, their real-world perception remains limited by insufficient multimodal coverage. Most tools provide textual information, lacking integration of visual, audio, or video modalities, which hampers accurate understanding in real-world reasoning tasks. Recent efforts to enhance multimodal processing—such as OpenAI’s O3 “Thinking with Images” framework \cite{su2025thinking} and related works \cite{hu2024visual, huang2025visualtoolagent, chern2025thinking}—improve LLMs’ ability to acquire and leverage multimodal information. These developments enable more comprehensive perception of objective facts.
\subsection{Clash of Caps}

Reasoning traits associated with different “hats” often conflict. For instance, the blue hat demands systematic rigor, while the green hat encourages creative exploration; the black hat emphasizes critical reflection, whereas the yellow hat favors rapid judgment; the red hat guides reasoning by preferences, while the white hat prioritizes objective facts. Balancing these traits within a single system is challenging, as models must simultaneously manage rigor, creativity, reflectiveness, efficiency, and factual accuracy, yet current methods struggle with optimal multi-hat coordination. Therefore, designing effective multi-hat collaboration mechanisms is a promising direction for human-like reasoning systems.

\section{Conclusion}
\label{sec: conclusion}
In this survey, we offer the first comprehensive review that connects technical developments with human reasoning mechanisms by mapping diverse CoT fine-tuning approaches to the Six Thinking Hats theory. Then, our analysis suggests that future research should pursue a more holistic view of reasoning in LLMs, emphasizing balanced integration across planning, creativity, judgment, reflection, internalization, and factual grounding. To facilitate progress, we also curated datasets, summarized model performance, and released a continuously updated repository to track recent advances.


\bibliographystyle{IEEEtran} 
\bibliography{custom} 

\vfill

\end{document}